\documentclass{article}

\usepackage{PRIMEarxiv}

\usepackage[utf8]{inputenc} 
\usepackage[T1]{fontenc}    
\usepackage{hyperref}       
\usepackage{url}            
\usepackage{booktabs}       
\usepackage{amsfonts}       
\usepackage{nicefrac}       
\usepackage{microtype}      
\usepackage{lipsum}
\usepackage{fancyhdr}       
\usepackage{amsmath,amsfonts}
\usepackage{algorithmic}
\usepackage{algorithm}
\usepackage{array}
\usepackage{textcomp}
\usepackage{stfloats}
\usepackage{url}
\usepackage{verbatim}
\usepackage{graphicx}
\usepackage{cite}
\usepackage{xcolor}
\usepackage{multicol}
\usepackage{multirow}
\usepackage{amsmath}
\usepackage{amsfonts,amssymb}
\usepackage{bm}
\usepackage{amsthm}
\usepackage{booktabs}
\usepackage{algorithm}
\usepackage{bbding}
\usepackage{pifont}
\usepackage{wasysym}
\usepackage{amssymb}
\usepackage{algorithmic}
\usepackage{xcolor}
\usepackage{times}
\usepackage{epsfig}
\usepackage{graphicx}
\usepackage{amsmath}
\usepackage{subfigure}
\usepackage{float}
\usepackage{graphicx}
\usepackage{multicol}
\usepackage{multirow}
\usepackage{amssymb}
\usepackage{bm} 
\usepackage{color}
\graphicspath{{media/}}     

\newcommand{\HL}[1]{\textcolor[rgb]{1.00,0.00,0.00}{#1}}
\newcommand{\hl}[1]{\textcolor[rgb]{0.00,0.00,1.00}{#1}}

\title{The Cascaded Forward Algorithm for Neural Network Training}

\author{
	Gongpei Zhao, Tao Wang, Yidong Li, Yi Jin, Congyan Lang \\
	Beijing Jiaotong University \\
  \texttt{\{csgpzhao, twang, ydli, yjin, cylang\}@bjtu.edu.cn} \\
   \And
  Haibin Ling \\
  Stony Brook University \\
  \texttt{hling@cs.stonybrook.edu} \\
}

\begin{document}
\maketitle

\begin{abstract}
Backpropagation algorithm has played a significant role in the development of deep learning. However, there exist some limitations associated with this algorithm, such as getting stuck in local minima and experiencing vanishing/exploding gradients, which have led to questions about its biological plausibility. To address these limitations, alternative algorithms to backpropagation have been preliminarily explored, with the Forward-Forward (FF) algorithm being one of the most well-known. In this paper we propose a new learning framework for neural networks, namely \textbf{Ca}scaded \textbf{Fo}rward  (\textbf{CaFo}) algorithm, which does not rely on BP optimization as that in FF. Unlike FF, CaFo directly outputs label distributions at each cascaded block and waives the requirement of generating additional negative samples. Consequently, CaFO leads to a more efficient process at both training and testing stages. Moreover, in our CaFo framework each block can be trained independently, allowing easy deployment to parallel acceleration systems. The proposed method is evaluated on four public image classification benchmarks, and the experimental results illustrate significant improvement in prediction accuracy in comparison with recently proposed baselines. Our
code is available at: \url{https://github.com/Graph-ZKY/CaFo}.
\end{abstract}

\section{Introduction}
\begin{figure}[t]
	\centering
	\subfigure[]{
		\begin{minipage}[b]{.8\textwidth}
			\includegraphics[width=1\textwidth]{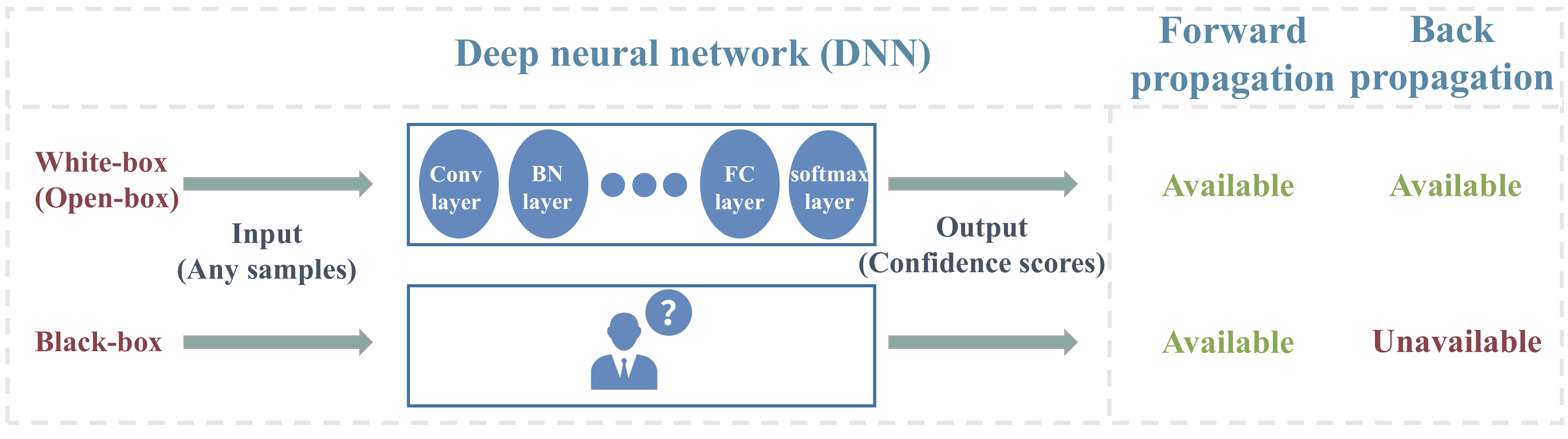}
		\end{minipage}
		
	}
	\subfigure[]{
		\begin{minipage}[b]{.8\textwidth}
			\includegraphics[width=1\textwidth]{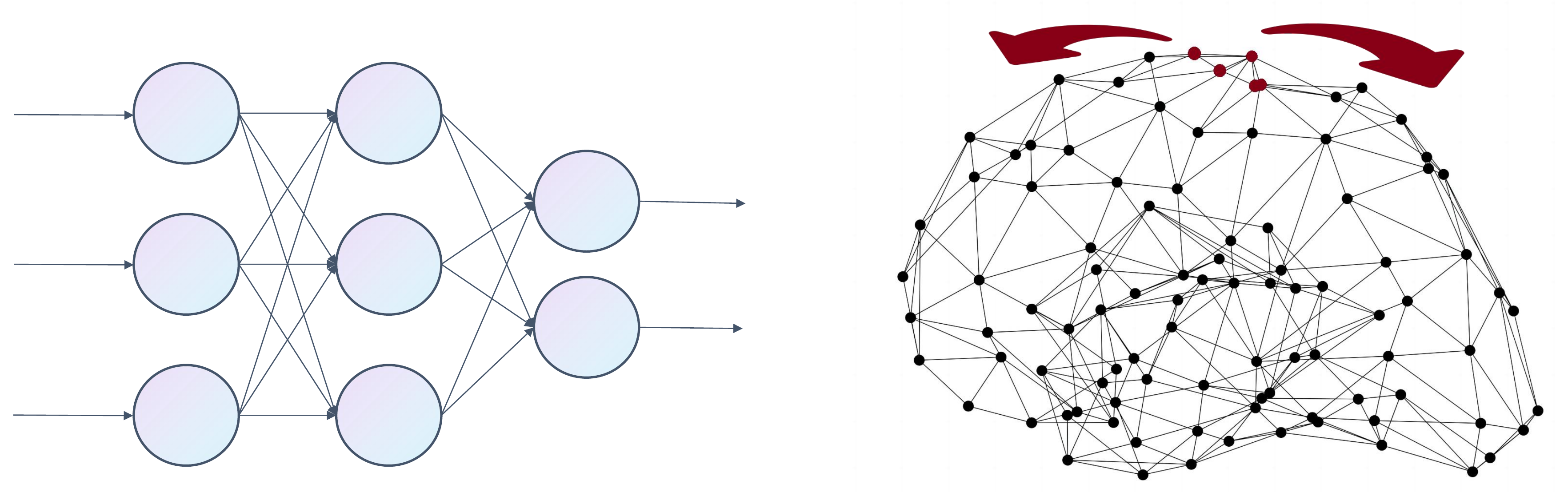}
		\end{minipage}
		
	}
	\caption{Intuitive implausibilities of backpropagation algorithm: (a) BP requires access to the internal structure of the neural network, including the activation functions and the number of layer. This is not always possible in black box systems, where the internal workings of the system are not transparent. (b) The active cortex (indicated with red) processes information in multiple directions, while in neural networks, information typically flows in a single direction to optimize a specific objective.}
	\label{fig:motivation}
	
\end{figure}

Backpropagation (BP) algorithm~\cite{rumelhart1986learning} is a powerful technique that has proven to be effective for training deep neural networks on a wide range of tasks, including image classification~\cite{tang2022decision}, object detection~\cite{zhao2019object}, machine translation~\cite{zhang2020neural}, text summarization~\cite{yang2020hierarchical} and graph learning~\cite{zhao2022neighborhood}. The algorithm's ability of adjusting the weights based on the error between the prediction and the ground truth allows the network to learn and improve over time, making it a cornerstone of deep learning and artificial intelligence. Nevertheless, despite its effectiveness, BP suffers from several limitations in practical applications. These limitations include the problems of local minima~\cite{baldi1989neural}, vanishing/exploding gradients~\cite{hanin2018neural}, overfitting~\cite{lawrence2000overfitting}, slower convergence and non-convex optimization~\cite{dauphin2014identifying}, which may negatively impact the training process. Additionally, BP relies on complete understanding of the computations performed during the forward pass in order to correctly calculate the derivatives. This characteristic makes it difficult to be generalized to the black-box systems where the internal working processes are not transparent, as shown in Figure~\ref{fig:motivation} (a). For biological plausibility, it seems that backpropagation remains implausible as a model of how cortex learns, despite considerable effort to invent ways in which it could be implemented by real neurons. For example, as shown in Figure~\ref{fig:motivation} (b), the connections between different cortical areas do not mirror the bottom-up connections of backpropagation-based deep learning models. Instead, they go in loops, in which neural signals traverse several cortical layers and then return to earlier areas. The bottom-up connections of backpropagation leads to the detachment of learning and inference, in which the training algorithm must stop inference to perform backpropagation in order to adjust the weights of a neural network. In contrast, the brain receives a constant stream of information, and the perceptual system needs to perform inference and learning in real time without stopping to perform backpropagation.

Due to the aforementioned limitations and the apparent difference in mechanisms between BP neural network and real cortical neurons as shown in Figure~\ref{fig:motivation}, some researchers~\cite{lillicrap2016random,nokland2016direct,frenkel2021learning,dellaferrera2022error,hinton2022forward} have raised concern about the biological plausibility of BP, questioning whether whether it has some other way of getting the gradients needed to adjust the weights on connections. This has prompted researchers to search for alternate algorithms to train deep neural networks. One of the most recent alternatives is the Forward-Forward (FF) algorithm~\cite{hinton2022forward}, which replaces the forward and backward passes of BP with two forward passes that work on positive and negative data with opposite optimization objectives. FF initiates preliminary and inspiring investigations in this field, demonstrating its potential as an alternative to BP. This opens up significant opportunities for further research and in-depth exploration.

Thus motivated, in this paper we present a flexible and effective learning framework for neural networks, namely \textbf{Ca}scaded \textbf{Fo}rward (\textbf{CaFo}) algorithm, which offers a promising alternative to the traditional BP algorithm. Our CaFo framework consists of multiple cascaded neural blocks, with a layer-wise predictor attached to each block. Each neural block is composed of one or more convolutional layers, pooling layers, normalization layers, activation layers, etc., of which the aim is to project the input feature map into a new feature space. All the neural blocks are concatenated in sequence and pre-trained according to a non-backpropagation method. Each layer-wise predictor takes the feature map computed by the corresponding pre-trained neural block as input and outputs the prediction of the task, and it is trained independently without backpropagation according to the errors between the output and the ground truth. During the test period, the final prediction is a combination of the predictions from all predictors.

Roughly speaking, instead of performing backpropagation using the chain rule, the CaFo network performs only forward pass to directly estimate the prediction errors and updates the parameters \textit{in situ}. Overall, our algorithm is a step forward on the basis of the FF algorithm and brings several advantages. Firstly, it eliminates the necessity for negative sampling, thereby reducing the impact of sampling quality on model performance and increasing stability. Secondly, our method directly outputs the prediction of multi-class distribution rather than a simple goodness estimation, and it is thus more suitable for multi-class prediction tasks. Finally, the pre-trained neural blocks enable independent training of the attached predictor without dependencies on the training outcomes of previous blocks. This design not only enhances portability and flexibility compared with FF but also facilitates straightforward deployment into parallel acceleration systems in certain scenarios.

For evaluation we test our algorithm on four public image classification benchmarks, in comparison with FF and other non-backpropagation algorithms. The experimental results show that our CaFo algorithm exhibits effectiveness and efficiency on image classification tasks, and outperforms all compared baselines by a remarkable margin.

In summary, with the proposed learning procedure for neural networks, we make contribution in three folds:

\begin{itemize}
	\item We propose CaFo, a novel neural network training procedure that offers significant improvements in image classification task compared with FF and other state-of-the-art non-BP approaches.
	\item We develop distinct training strategies for the two core components of CaFo (i.e., neural blocks and predictors), respectively, to address the diverse requirements of both training efficiency and prediction effectiveness.
	
	\item Extensive experiments are conducted, and the experimental results demonstrate the superiority of our method against the state-of-the-art approaches.
\end{itemize}

This article is structured as follows: In Section II, we begin with a concise review of the well-known backpropagation algorithm and some representative non-backpropagation (non-BP) algorithms. Section III presents our proposed method, CaFo, encompassing notations, a pipeline overview, and the detailed introduction of neural block training, predictor training, and the inference process. In Section IV, we delve into discussions regarding the biological plausibility of CaFo and its relationship to some classic algorithms. Subsequently, in Section V, we present detailed experimental results and performance analyses obtained from real-world datasets. Finally, in Section VI, we present future work and provide concluding remarks.

\section{Related Work}

\subsection{Backpropagation Algorithm}

Backpropagation algorithm is a widely used training algorithm for deep neural networks, the goal of which is to optimize the parameters of the networks by minimizing the prediction error between the network's output and the ground truth. BP uses the gradient descent algorithm for optimization. The gradient of the loss function is calculated using the chain rule~\cite{clark1997constructing} of differentiation, where the error is propagated backwards through the network and the parameters are updated according to the negative gradient. 

BP is favored for its ease of implementation, computational efficiency, and effectiveness on a variety of problems. However, it also suffers from certain limitations that have been deeply studied~\cite{srivastava2014dropout,zhong2020random,prechelt2012early,glorot2010understanding,zhang2019gradient,xu2020reluplex,he2016deep,ioffe2015batch}. For the overfitting problems in deep neural network, some studies propose regularization tricks such as L1, L2 regularization~\cite{bektacs2010comparison} and dropout~\cite{srivastava2014dropout} to enhance the generalizability of models. Other works propose to solve this problem by data augmentation~\cite{zhong2020random} and training tricks (e.g. early stopping~\cite{prechelt2012early}). For the gradient problems during training, there are several techniques to address the vanishing and exploding gradients, including weight initialization~\cite{glorot2010understanding}, gradient clipping~\cite{zhang2019gradient}, non-linear activation functions~\cite{xu2020reluplex}, skip connections~\cite{he2016deep}, and normalization techniques~\cite{ioffe2015batch}. To address the problems of local minima and slow convergence, mini-batch training is encouraged, and some ingenious optimization algorithms for deep neural network is designed, such as Adam~\cite{kingma2014adam} and RMSProp~\cite{ruder2016overview}. 

Despite these efforts, problems with backpropagation still occur stochastically, highlighting the need for an in-depth understanding of deep learning and experience in model tuning. Due to the limitations mentioned above, some recent researches~\cite{hinton2022forward,ororbia2023predictive} have even raised doubts about the biological plausibility of BP.

\begin{figure*}[t]
	\begin{center}
		\includegraphics[width=0.98\textwidth]{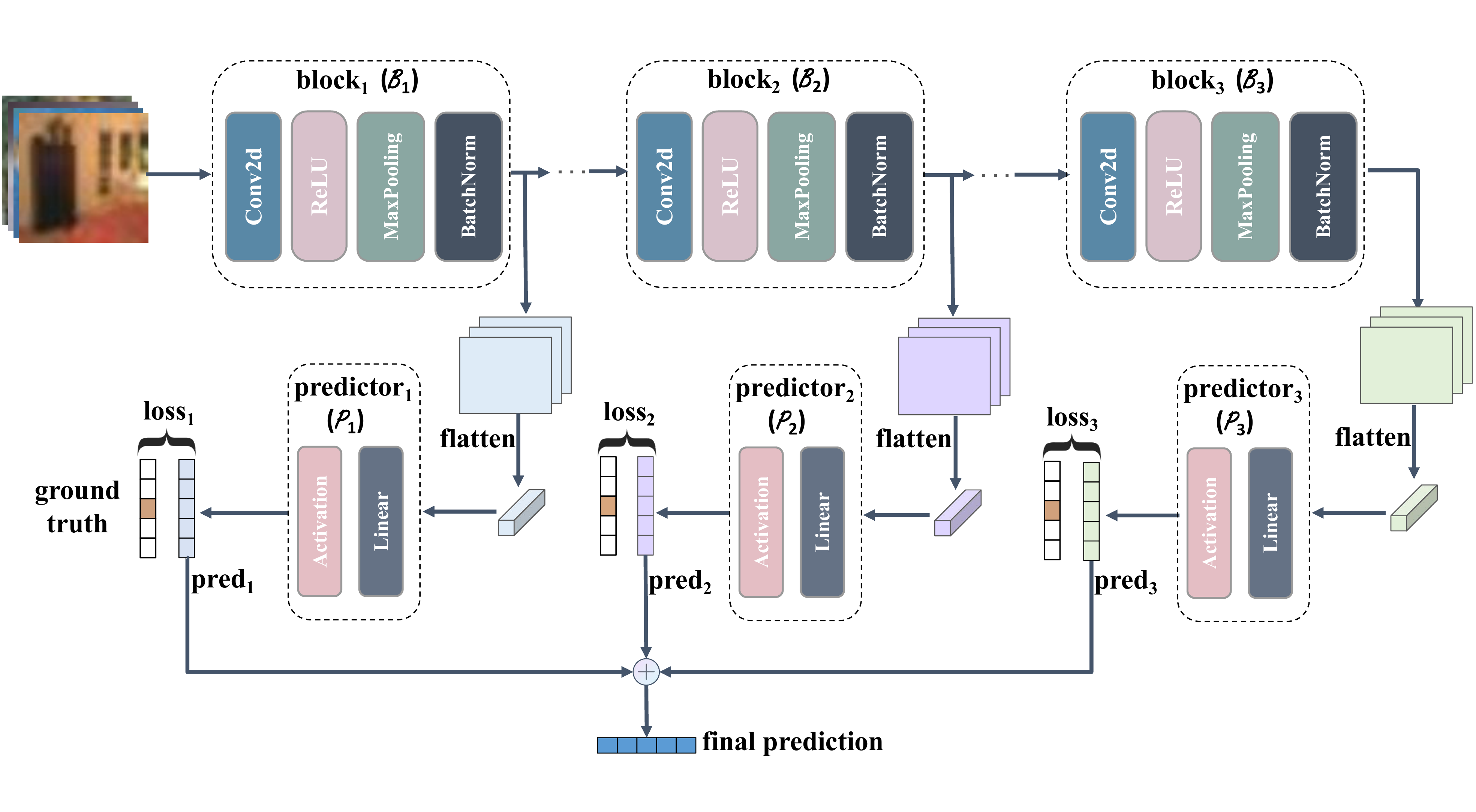}
	\end{center}
	\caption{Overall architecture of the CaFo network. Two main components are cascaded in the network: (1) a neural block that extracts representation from source data (e.g., images) at a specific scale, and (2) a predictor that takes the intermediate representation extracted by the corresponding block as input and outputs the prediction results.} 
\label{fig:framework}
\end{figure*}

\subsection{Non-backpropagation Algorithm}

The biological implausibility of the BP mainly lies in the problems of weight transport~\cite{grossberg1987competitive}, non-locality, freezing of neural activity, and update locking~\cite{jaderberg2017decoupled}. The feedback alignment algorithm (FA)~\cite{lillicrap2016random} solves weight transport problem and demonstrates that using fixed random weights in the feedback pathway allows conveying useful error gradient information. Following this pipeline, the DFA~\cite{nokland2016direct}, DRTP~\cite{frenkel2021learning} and PEPITA~\cite{dellaferrera2022error} algorithms are proposed in sequence, which partially address the remaining three problems by introducing non-backpropagation updating algorithms for training. The most recent concern raised by Hinton~\cite{hinton2022forward} about the biological plausibility of the BP has spurred the introduction of the FF algorithm as an alternative to better align with neurophysiological findings. FF replaces the forward and backward passes of BP with two forward passes that work on positive and negative data with opposite optimization objectives. FF has been validated to be an effective alternative to BP in some tasks, due to its simplicity and flexibility for network architecture and its ability to handle non-differentiable components. Nevertheless, some potential limitations of FF make it hard to be integrated into existing learning systems, including the high computational demanding of two forward passes, the high dependence on positive and negative sampling strategies, and the rough approximation of optimization objective. 

Our CaFo framework follows this pipeline and works without backpropagation. Moreover, it eliminates the need for negative sampling and updates the parameters directly according to the errors between the predicted output and the ground truth. It simplifies both training and testing processes, and gains significant improvement in prediction accuracy.

\section{The Proposed Method}\label{sec:method}

In this section, we describe the details of  the proposed CaFo framework. Following the work of FF, we apply this method to the task of image classification, a well-established challenge in computer vision. To ensure a clear and consistent presentation throughout the paper, we firstly provide an overview of the notations and mathematical symbols used in our analysis. This is followed by an elaboration of the overall architecture of the CaFo framework. Finally we describe the training and inference processes of the proposed approach.

\subsection{Notations}
To ensure clarity and consistency in our presentation, we provide a brief overview of the notations we used in this paper. The scalars used in our analysis are represented by italic lowercase letters (e.g., \textit{m}, \textit{n}), while the vectors used in our method are represented by bold lowercase letters (e.g., \textbf{x}, \textbf{y}). All vectors in this paper are assumed to be column vectors. Bold uppercase letters (e.g., \textbf{W}, \textbf{Z}) are used to represent matrices. Furthermore, the subscripts and superscripts in our notations represent the indices and exponents, respectively. In the rest of the paper, we will provide more detailed definitions and explanations of the notations as needed.

\subsection{Pipeline Overview}

The pipeline of the proposed CaFo framework is depicted in Figure~\ref{fig:framework}, which consists of multiple repeated stacks of two key components:

\textbf{Neural Block:} The neural block, abbreviated as \textit{block} for convenience in the rest of the paper, is comprised of multiple element units, such as convolutional layer, pooling layer, normalization layer, activation layer, etc. Its architecture depends on the specific task. For image classification in our experiments, it consists of a convolutional layer followed by a ReLU activation function, a max-pooling layer and a batch normalization layer.  Multiple blocks with the same or different architectures are cascaded to extract intermediate image representations at different scales. It is worth noting that the block can be either parametric or non-parametric, which provides flexibility in its construction.

\textbf{Predictor:} As illustrated in Figure~\ref{fig:framework}, each block is equipped with a layer-wise predictor that consists of a fully connected layer followed by an activation function (e.g., softmax) if necessary. The predictor takes the intermediate representation extracted by the block as input and outputs a prediction result for the task. Since there is only one fully connected layer, the predictor can be trained \textit{in situ} without backpropagation.

As described below, each block is trained separately, independent of the predictors in our network.


\subsection{Training of Neural Block}

Inspired by DFA~\cite{nokland2016direct}, we employ a non-BP approach to train the blocks. Firstly, we ignore the predictors and concatenate all the blocks in sequence. An additional output layer is introduced, performing a linear transformation followed by sigmoid activation after the last block. We firstly introduce the training strategy for the fully connected layer and subsequently extend it to cover the convolutional layer.
\paragraph{Fully connected layer}Let $(\text{\textbf{x}},\text{\textbf{y}})$ be the input-output vectors of a sample we want the network to learn. For simplicity, we assume that the network has only two blocks and omit normalization layers. Denoting that $\text{\textbf{V}}_{i}$ and $\text{\textbf{b}}_{i}$ the weights and biases for the units in hidden layer $i$, the forward pass in the network is calculated as:
\begin{gather}
	block_{1}: \text{\textbf{a}}_{1}=(\text{\textbf{V}}_{1} \text{\textbf{x}}) + \text{\textbf{b}}_{1}, \ \ \text{\textbf{h}}_{1}=f(\text{\textbf{a}}_{1}), \\
	block_{2}: \text{\textbf{a}}_{2}=(\text{\textbf{V}}_{2}\text{\textbf{h}}_{1}) + \text{\textbf{b}}_{2}, \ \ \text{\textbf{h}}_{2}=f(\text{\textbf{a}}_{2}), \\
	output\_layer: \text{\textbf{a}}_{y}=(\text{\textbf{V}}_{3}\text{\textbf{h}}_{2})+\text{\textbf{b}}_{3}, \ \ \hat{\text{\textbf{y}}}={\rm sigmoid}(\text{\textbf{a}}_{y}),
\end{gather}
where $f(.)$ the non-linearity used in hidden layers.

We update the parameters of blocks according to the classification error: $\text{\textbf{e}}=\hat{\text{\textbf{y}}}-\text{\textbf{y}}$. For the linear output layer, the weight and bias updates are calculated as:
\begin{gather}
	\text{\textbf{V}}_{3} \leftarrow  \text{\textbf{V}}_{3}-\eta\Delta\text{\textbf{V}}_{3}, \ \ \Delta\text{\textbf{V}}_{3}=-\text{\textbf{e}} \text{\textbf{h}}_{2}^{T}, \\
	\text{\textbf{b}}_{3} \leftarrow  \text{\textbf{b}}_{3}-\eta\Delta\text{\textbf{b}}_{3}, \ \ \Delta\text{\textbf{b}}_{3}=-\text{\textbf{e}},
\end{gather}
where $\eta$ is the learning rate. For fully connected layer, in BP the changes of weights (i.e., $\Delta\text{\textbf{V}}_{i}$) and biases (i.e., $\Delta\text{\textbf{b}}_{i}$) depend on the value of $\text{\textbf{V}}_{i+1}$ and the error gradient $\Delta \text{\textbf{h}}_{i}$. As for non-BP framework, we follow DFA to use a fixed random feedback matrix $\text{\textbf{B}}_i$ for each fully connected layer as follows:
\begin{align}\label{eq:ce11}
	\Delta \text{\textbf{h}}_{i}=\text{\textbf{B}}_{i}\text{\textbf{e}}.
\end{align}
By employing this approach, the error gradient for each fully connected layer can be calculated in parallel, waiving the need for a backward pass of the error gradient. 

\paragraph{Covolutional layer}Let $(\text{\textbf{X}},\text{\textbf{y}})$ be the input-output tensors of a sample we want the network to learn. For $i$-th block we have:
\begin{align}\label{eq:conv}
	block_{i}: \text{\textbf{a}}_{i}=(\text{\textbf{V}}_{i} \ast \text{\textbf{h}}_{i-1}) + \text{\textbf{b}}_{i}, \ \ \text{\textbf{h}}_{i}=f(\text{\textbf{a}}_{i}), 
\end{align}
where $\ast$ denotes 2-D convolution operation, and $f(.)$ the non-linearity used in hidden layers. We initialize $\text{\textbf{h}}_{0}$ as $\text{\textbf{X}}$ for the first block, and for the sake of simplicity, we set the stride to one. The error gradient for each layer is calculated as
\begin{align}\label{eq:conv1}
	\Delta \text{\textbf{h}}_{i-1}=\Delta \text{\textbf{a}}_{i} \ast \textit{ROT180}(\text{\textbf{V}}_{i}),
\end{align}
where $ROT180(.)$ denotes a 180-degree rotation. Since $\Delta \text{\textbf{a}}_{i}$ is associated with the error term $\text{\textbf{e}}=\hat{\text{\textbf{y}}}-\text{\textbf{y}}$, we represent this relationship as $\Delta \text{\textbf{a}}_{i}=g_{i}(\text{\textbf{e}})$. Following the approach of DFA, we employ a fixed random feedback matrix for each convolutional layer, which is outlined as follows:
\begin{align}\label{eq:conv2}
	\Delta \text{\textbf{h}}_{i-1}=g_{i}(\text{\textbf{e}}) \ast \textit{ROT180}(\text{\textbf{B}}_{i}).
\end{align}

For computational simplicity, we further employ Eq.~\ref{eq:ce11} to approximate Eq.~\ref{eq:conv2} during the training of convolutional layers.

Noting that, for simplicity the parameters of all blocks can be randomly parameterized without training.
The experimental results in Tables~\ref{tab:smalldata} and \ref{tab:largedata} demonstrate that, despite simplicity, the blocks parameterized with Kaiming uniform~\cite{he2015delving} can produce discriminative representations for prediction, achieving better performance in comparison with baseline algorithms. Moreover, blocks with finely trained parameters excel in extracting intermediate features and enhance the model performance.

\subsection{Training of Predictor}

For each predictor, we optimize it by minimizing the errors between the prediction and the ground truth.  Different loss functions can be adopted in our framework, which lead to different optimization strategies and algorithms. In experiments we compare three different loss functions: mean square error loss (MSE), cross-entropy loss (CE) and sparsemax loss (SL)~\cite{martins2016softmax}. Here we describe the corresponding formulations and optimization algorithms respectively.

\paragraph{MSE loss} The optimization objective of the predictor can be expressed as follows:
\begin{align}\label{eq:mse}
	\arg\min\limits_{\text{\textbf{W}}}\frac{1}{2m}\|\text{\textbf{HW}}-\text{\textbf{Y}}\|_{F}^{2},
\end{align}
where $m$ denotes the number of training samples, $\textbf{H}\in\mathbb{R}^{m\times d}$ the $d-$dimensional intermediate representation output by the block, $\textbf{Y} \in \mathbb{R}^{m \times c}$ the one-hot labels of training samples, $\textbf{W} \in \mathbb{R}^{d \times c}$ the parameters of predictor to be optimized, $\|\cdot\|_F$ the Frobenius norm, and $c$ indicates the number of classes. The close-form solution for Eq.~\ref{eq:mse} can be obtained by solving the following equation for $\textbf{W}$:
\begin{align}\label{eq:mse1}
	\frac{\partial}{\partial \bf{W}}(\frac{1}{2m}\|\textbf{HW}-\textbf{Y}\|_{F}^{2})=\frac{1}{m}\textbf{H}^{T}(\textbf{HW}-\textbf{Y})=0,
\end{align}
and the solution of Eq.~\ref{eq:mse} is:
\begin{align}\label{eq:mse2}
	\textbf{W}=(\textbf{H}^{T}\textbf{H})^{-1}\textbf{H}^{T}\textbf{Y}.
\end{align}

\paragraph{Cross-entropy loss} The optimization objective of the predictor is written as:
\begin{gather}
	\arg\min \limits_{\textbf{W}}-\frac{1}{m}\sum_{i=1}^{m}\sum_{j=1}^{c}\textbf{Y}_{i,j} \cdot \ln \textbf{P}_{i,j}, \label{eq:ce}\\
	\textbf{P}_{i,j}=\frac{\exp(\textbf{H}_{i,:}\textbf{W}_{:,j})}{\sum_{k=1}^{c}\exp(\textbf{H}_{i,:}\textbf{W}_{:,k})},\label{eq:ce1}
\end{gather}
%
%
where $\textbf{Y}_{i,j}$ is the value at $i$-th row and $j$-th column of $\textbf{Y}$, $\textbf{H}_{i,:}$ and $\textbf{W}_{:,j}$ respectively denote the $i$-th row of $\textbf{H}$ and the $j$-th row of $\textbf{W}$, and $\textbf{P}_{i,j}$ the corresponding output of softmax. 

As the close-form solution for Eq.~\ref{eq:ce} is unavailable, we use gradient descent algorithm to optimize it. To do this, we need to calculate the gradient for $\textbf{W}$ according to the Jacobian matrix, and set a fixed step size to optimize Eq.~\ref{eq:ce} iteratively. The Jacobian matrix is calculated by $\textbf{J}=\textbf{P}-\textbf{Y}$,
and the gradient for $\textbf{W}$ is computed as:
\begin{align}\label{eq:ce3}
	\frac{1}{m}\sum_{i=1}^{m}[\textbf{H}_{i,:}]^{T}\otimes \textbf{J}_{i,:},
\end{align}
where $\otimes$ denotes the Kronecker product.


\paragraph{Sparsemax loss} Different than softmax, the sparsemax function~\cite{martins2016softmax} produces sparse output probabilities by enforcing a constraint that the output confidence vector has at most a certain number of non-zero elements. It encourages the model to only assign high probabilities to the most relevant classes, while setting all other probabilities to zero. The sparsemax function achieves this by projecting the input vector onto a simplex, which is a convex polytope whose vertices lie on the coordinate axes. The formulaic expression for sparsemax is as follows:
\begin{align}\label{eq:ce4}
	{\rm sparsemax}(\textbf{z}):=\arg\min\limits_{\textbf{p}\in \bigtriangleup^{c-1}}\|\textbf{p}-\textbf{z}\|^{2},
\end{align}
where $\bigtriangleup^{c-1}:=\{\textbf{p} \in \mathbb{R}^{c}|\textbf{1}^{T}\textbf{p}=1,\textbf{p} \ge \textbf{0}\}$ is the $(c-1)$-dimensional simplex. The resulting projection is unique and can be computed efficiently using a sorting algorithm~\cite{martins2016softmax}. 

We also use the gradient descent algorithm to approximate the sparsemax loss optimization objective:
\begin{gather}
	\arg\min\limits_{\textbf{W}}\frac{1}{m}\sum_{\textbf{z} \in \textbf{Z}}^{}\sum_{k=1}^{c}L_{\rm sparse}(\textbf{z};k),\label{eq:sl} \\
	L_{\rm sparse}(\textbf{z};k)=-\textbf{z}_{k}+\frac{1}{2}\sum_{j \in S(\textbf{z})}(\textbf{z}_{j}^{2}-\tau^{2}(\textbf{z}))+\frac{1}{2}, \label{eq:sl1}
\end{gather}
%
%
where $\textbf{z} \in \mathbb{R}^{c}$ is the output logits, $\textbf{Z}=[\textbf{z}_{1}; \textbf{z}_{2}; ...; \textbf{z}_{m}]^{T} \in \mathbb{R}^{m \times c}$ the set of output logits of $m$ training samples, $S(\textbf{z})$ the support of ${\rm sparsemax}(\textbf{z})$, and $\tau(\textbf{z})=\frac{(\sum_{j \in S(\textbf{z})}\textbf{z}_{i})-1}{|S(\textbf{z})|}$. The Jacobian matrix is calculated by:
\begin{align}\label{eq:sl2}
	\text{\textbf{J}}={\rm sparsemax}(\text{\textbf{Z}})-\text{\textbf{Y}},
\end{align}
\noindent and the gradient for \textbf{W} is computed using the same formula as Eq.~\ref{eq:ce3}.

\subsection{Inference Process}

In the inference phase each predictor outputs a prediction and the final prediction is determined by combining all the individual ones. In classification tasks this is typically done by summing the predictions of all the predictors and choosing the index of the maximum value as the predicted class.


The detailed optimization algorithm is shown in Algorithm~\ref{alg:algorithm1}. During the training stage, the model generates the prediction block by block at step 3 and step 4, where $\textbf{h}_{j}^{i}\in \mathbb{R}^{d} $ and $\tilde{\textbf{y}}_{j}^{i} \in \mathbb{R}^{c}$ are the intermediate representation and prediction of $i$-th block for $j$-th training sample. 
At step 5, the predictors are updated based on a specific loss function (e.g., Eqs.~\ref{eq:mse}, ~\ref{eq:ce}, and ~\ref{eq:sl}), using the corresponding optimization strategy. 
The predictors consist of a fully connected layer for MSE, and an additional activation layer is followed  for CE and SL. During the inference stage, each predictor outputs a prediction for the test set at step 3 and step 4, where $\hat{\textbf{y}}_{j}^{i} \in \mathbb{R}^{c}$ is the prediction of $i$-th block for the $j$-th test sample. Then the predictions of each block (i.e., $\hat{\text{\textbf{Y}}}^{i} \in \mathbb{R}^{n \times c}$) are summed to derive the final prediction $\hat{\textbf{\textbf{Y}}} \in \mathbb{R}^{n \times c}$. The predicted class for $j$-th test samples is then calculated at step 7 as: $\hat{y}_{j}=\arg\max\limits_{k} \hat{\textbf{\textbf{Y}}}_{j,k}$.

\begin{algorithm}[!t]
	\caption{The CaFo Algorithm}
	\label{alg:algorithm1}
	\textbf{Input}: \\
	$\text{\textbf{X}}=[\text{\textbf{x}}_{1}; \text{\textbf{x}}_{2}; ...; \text{\textbf{x}}_{m}]^{T}$: m samples for training, \\
	$\hat{\text{\textbf{X}}}=[\hat{\text{\textbf{x}}}_{1}; \hat{\text{\textbf{x}}}_{2}; ...; \hat{\text{\textbf{x}}}_{n}]^{T}$: n samples for inference, \\
	$\text{\textbf{Y}}=[\text{\textbf{y}}_{1}; \text{\textbf{y}}_{2}; ...; \text{\textbf{y}}_{m}]^{T}$: one-hot labels of training samples, \\
	$\mathcal{B}_{i}$: the $i$-th block, \\
	$\mathcal{P}_{i}$: the predictor of $i$-th block. \vspace{1mm}\\
	\textbf{Manual factors}: \\
	$r$: the number of blocks of CaFo \vspace{1mm}\\
	\textbf{Output}: $\bf{\hat{y}}$$=[\hat{y}_{1}, \hat{y}_{2}, ..., \hat{y}_{n}]^{T}$ \vspace{1mm}\\
	\textbf{Training:}
	\begin{algorithmic}[1] 
		\STATE ${\text{\textbf{H}}^{0}=[\text{\textbf{h}}^{0}_{1}; \text{\textbf{h}}^{0}_{2}; ...; \text{\textbf{h}}^{0}_{m}]^{T}=\text{\textbf{X}}}$
		\FOR{$i=1 \to r$}
		\STATE Get the intermediate representations of the $i$-th block: \\
		$\text{\textbf{H}}^{i}=[\text{\textbf{h}}^{i}_{1}; \text{\textbf{h}}^{i}_{2}; ...; \text{\textbf{h}}^{i}_{m}]^{T}=\mathcal{B}_{i}(\text{\textbf{H}}^{i-1})$  \\
		\STATE Get the predictions of the $i$-th block: \\
		$\tilde{\text{\textbf{Y}}}^{i}=[\tilde{\text{\textbf{y}}}^{i}_{1}; \tilde{\text{\textbf{y}}}^{i}_{2}; ...; \tilde{\text{\textbf{y}}}^{i}_{m}]^{T}=\mathcal{P}_{i}(\text{\textbf{H}}^{i})$  \\
		\STATE Optimizing $\mathcal{P}_{i}$ according to the errors between $\tilde{\text{\textbf{Y}}}^{i}$ and $\text{\textbf{Y}}$
		\ENDFOR
	\end{algorithmic}
	\vspace{1mm}
	\textbf{Inference:}
	\begin{algorithmic}[1] 
		\STATE $\hat{\text{\textbf{H}}}^{0}=[\hat{\text{\textbf{h}}}^{0}_{1}; \hat{\text{\textbf{h}}}^{0}_{2}; ...; \hat{\text{\textbf{h}}}^{0}_{n}]^{T}=\hat{\text{\textbf{X}}}$
		\FOR{$i=1 \to r$}
		\STATE Get the intermediate representations of the $i$-th block: \\
		$\hat{\text{\textbf{H}}}^{i}=[\hat{\text{\textbf{h}}}^{i}_{1}; \hat{\text{\textbf{h}}}^{i}_{2}; ...; \hat{\text{\textbf{h}}}^{i}_{n}]^{T}=\mathcal{B}_{i}(\hat{\text{\textbf{H}}}^{i-1})$  \\
		\STATE Get the predictions of the $i$-th block: \\
		$\hat{\text{\textbf{Y}}}^{i}=[\hat{\text{\textbf{y}}}^{i}_{1}; \hat{\text{\textbf{y}}}^{i}_{2}; ...; \hat{\text{\textbf{y}}}^{i}_{n}]^{T}=\mathcal{P}_{i}(\hat{\text{\textbf{H}}}^{i})$  \\
		\ENDFOR
		\STATE $\hat{\text{\textbf{Y}}}=\sum_{i=1}^{r}\hat{\text{\textbf{Y}}}^{i}$
		\STATE $\hat{\text{\textbf{y}}}=\arg\max\hat{\text{\textbf{Y}}}$
		\RETURN $\hat{\text{\textbf{y}}}$
	\end{algorithmic}
\end{algorithm}

\section{Discussion}

\begin{figure*}[t]
	\begin{center}
		
		\begin{minipage}[b]{1\textwidth}
			\centering
			\includegraphics[width=1\textwidth]{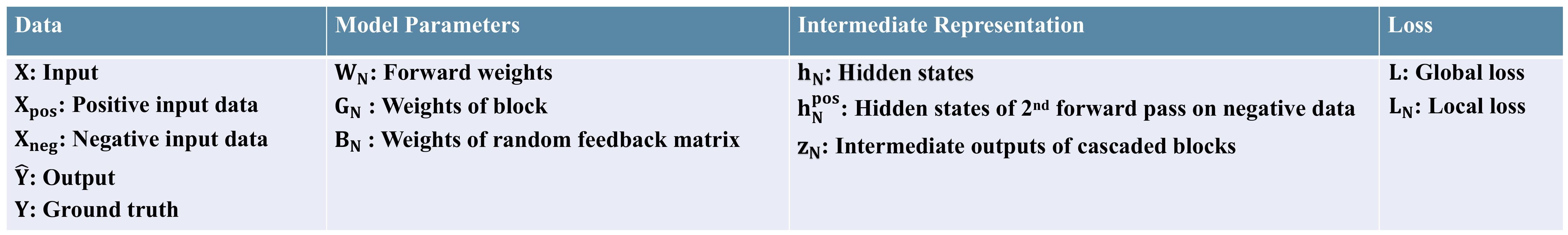}
		\end{minipage}
		
		\subfigure[BP]{
			\begin{minipage}[b]{0.30\textwidth}
				\centering
				\includegraphics[width=1\textwidth]{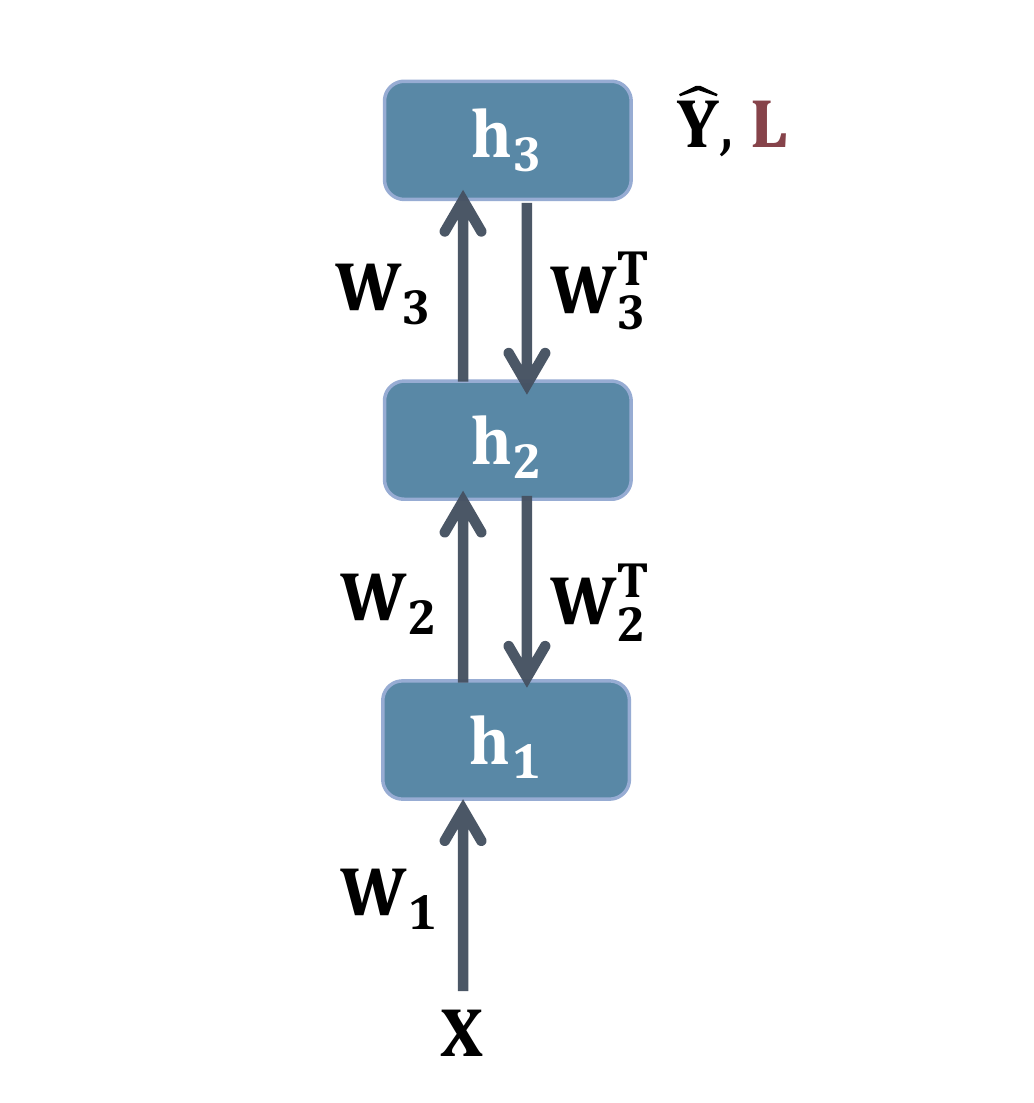}
			\end{minipage}
		}
		\subfigure[DFA]{
			\begin{minipage}[b]{0.30\textwidth}
				\centering
				\includegraphics[width=1\textwidth]{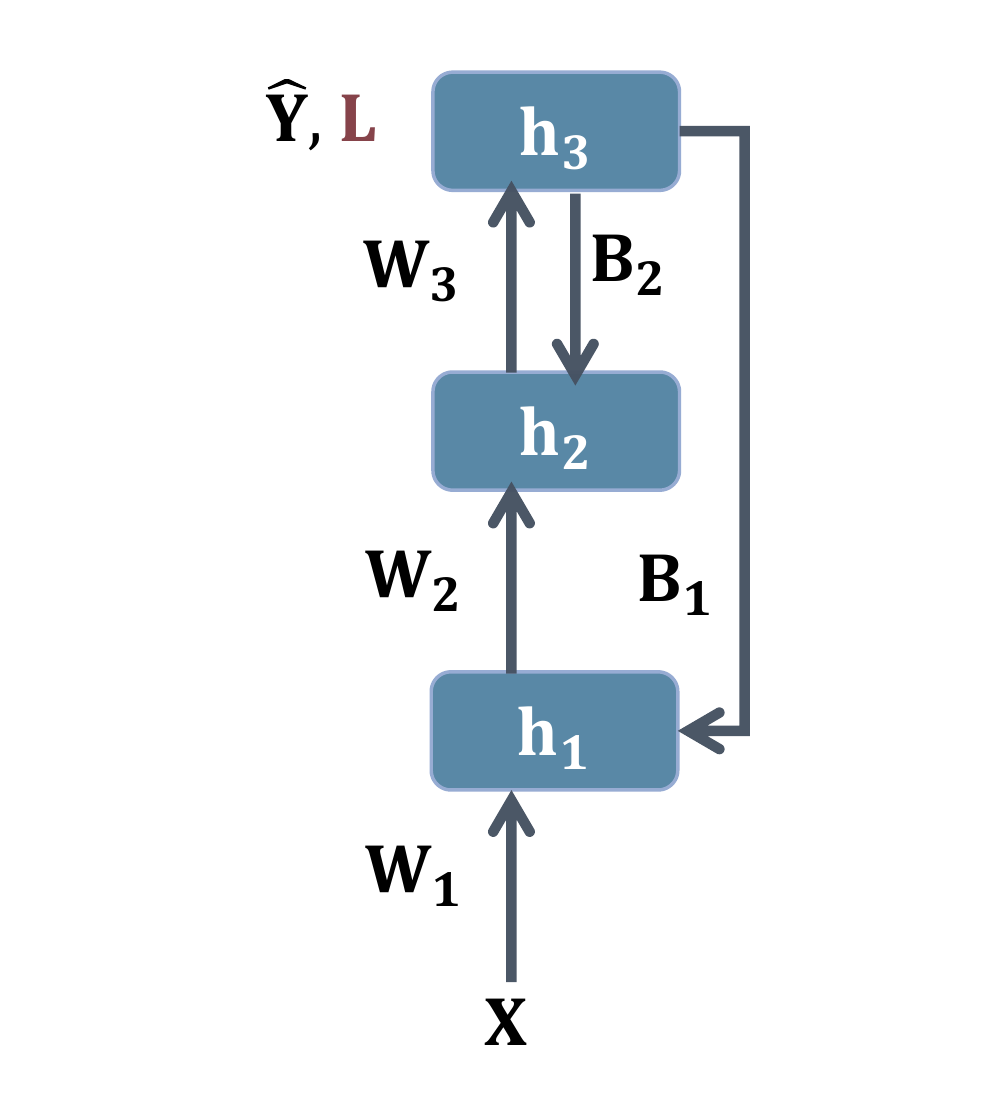}
			\end{minipage}
		}
		\subfigure[DRTP]{
			\begin{minipage}[b]{0.30\textwidth}
				\centering
				\includegraphics[width=1\textwidth]{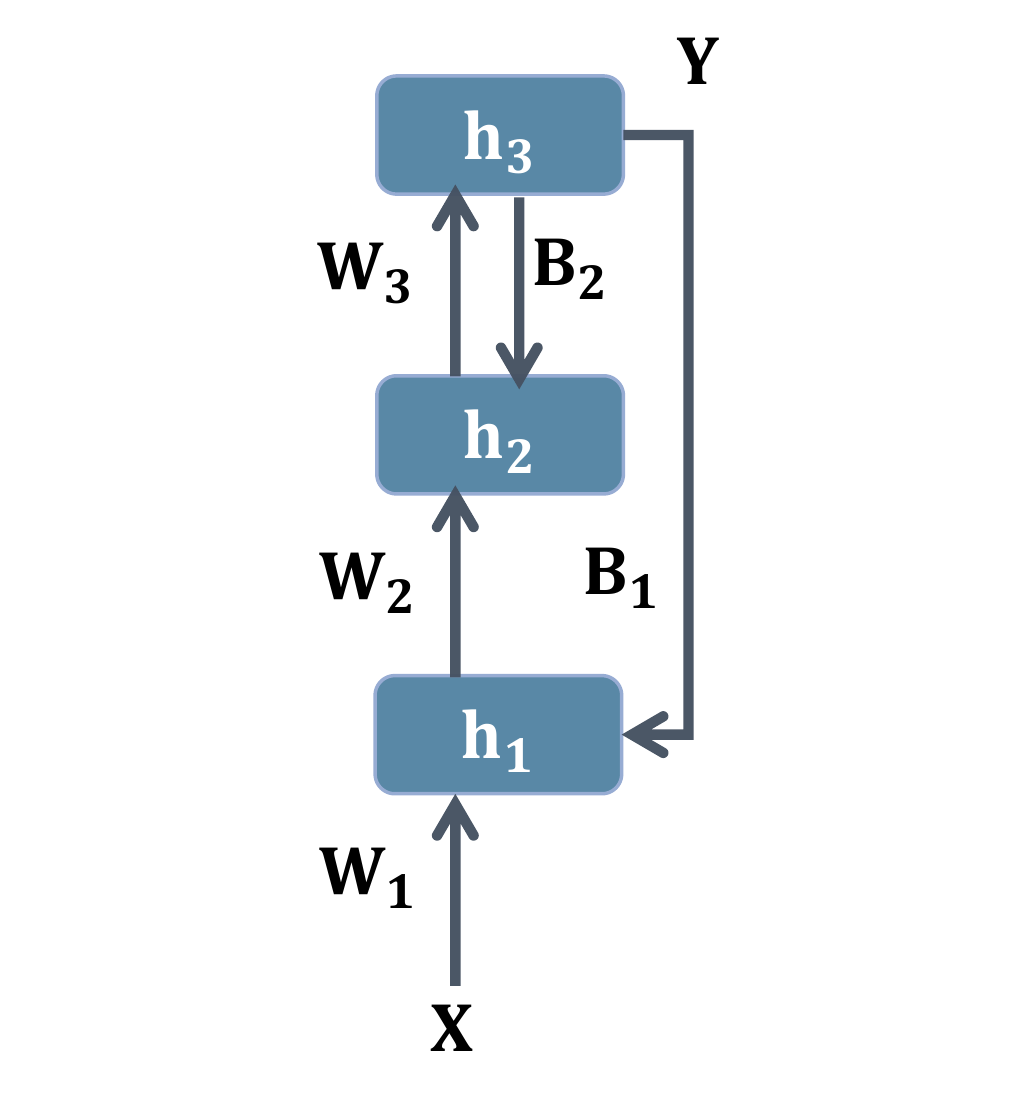}
			\end{minipage}
		}
		\newline
		\subfigure[PEPITA]{
			\begin{minipage}[b]{0.30\textwidth}
				\centering
				\includegraphics[width=1\textwidth]{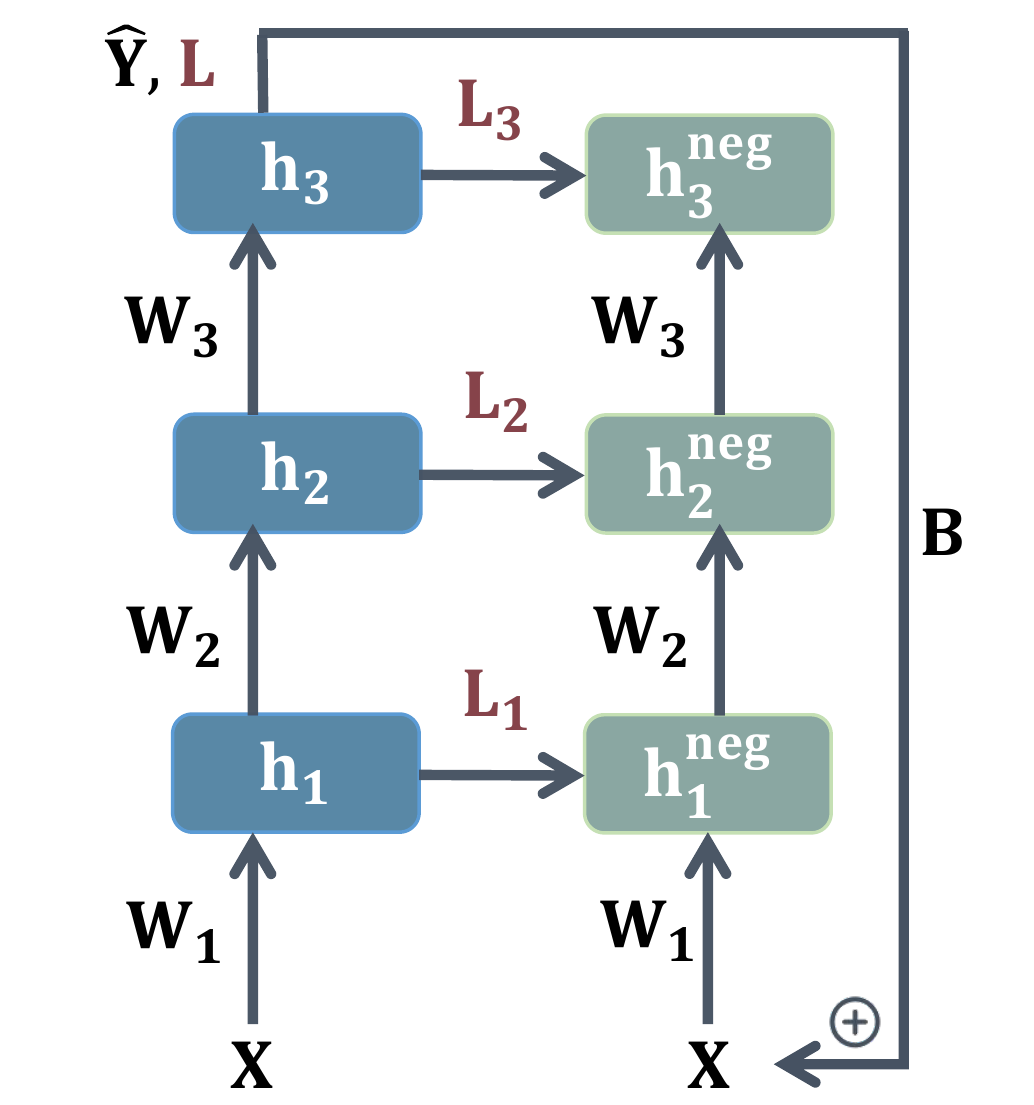}
			\end{minipage}
		}
		\subfigure[FF]{
			\begin{minipage}[b]{0.30\textwidth}
				\centering
				\includegraphics[width=1\textwidth]{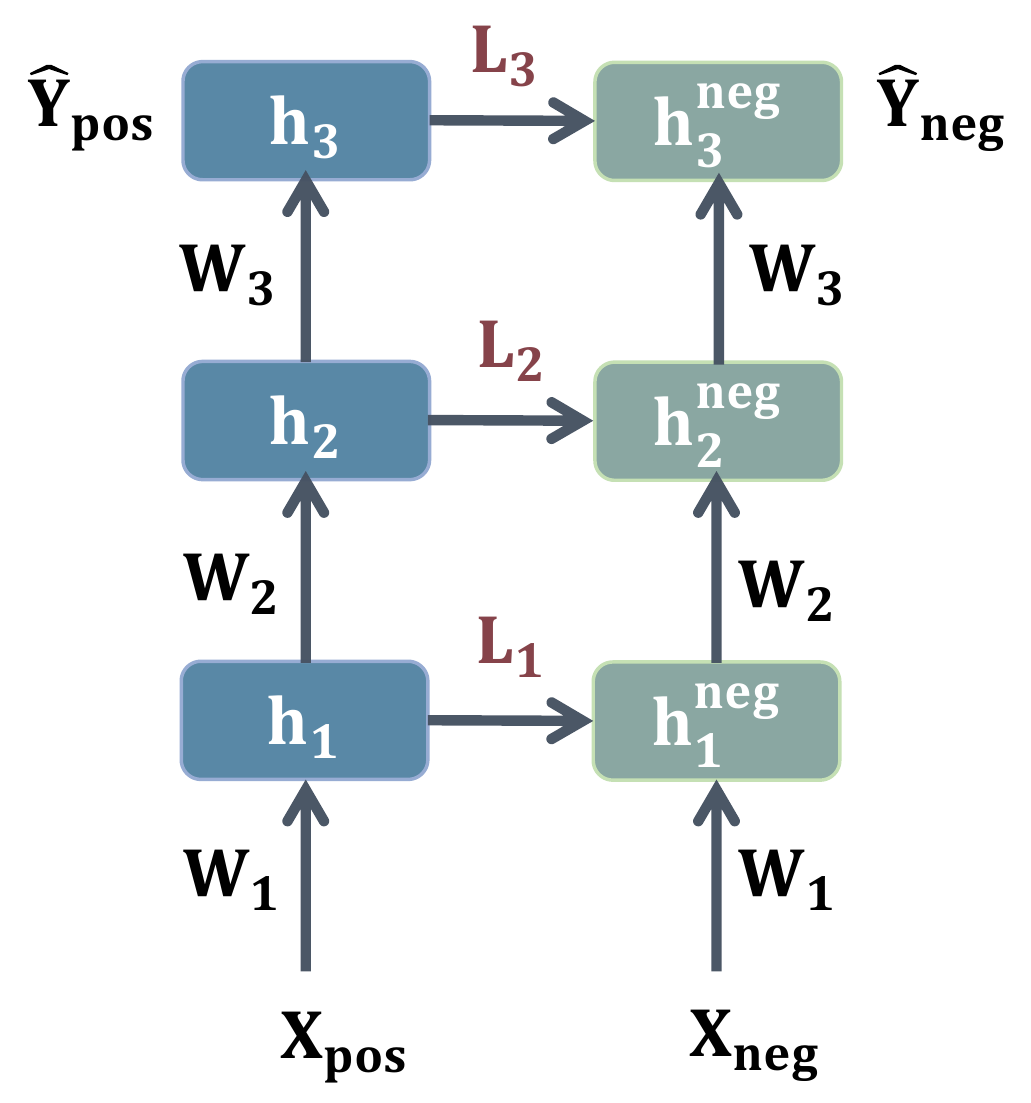}
			\end{minipage}
		}
		\subfigure[CaFo]{
			\begin{minipage}[b]{0.30\textwidth}
				\centering
				\includegraphics[width=1\textwidth]{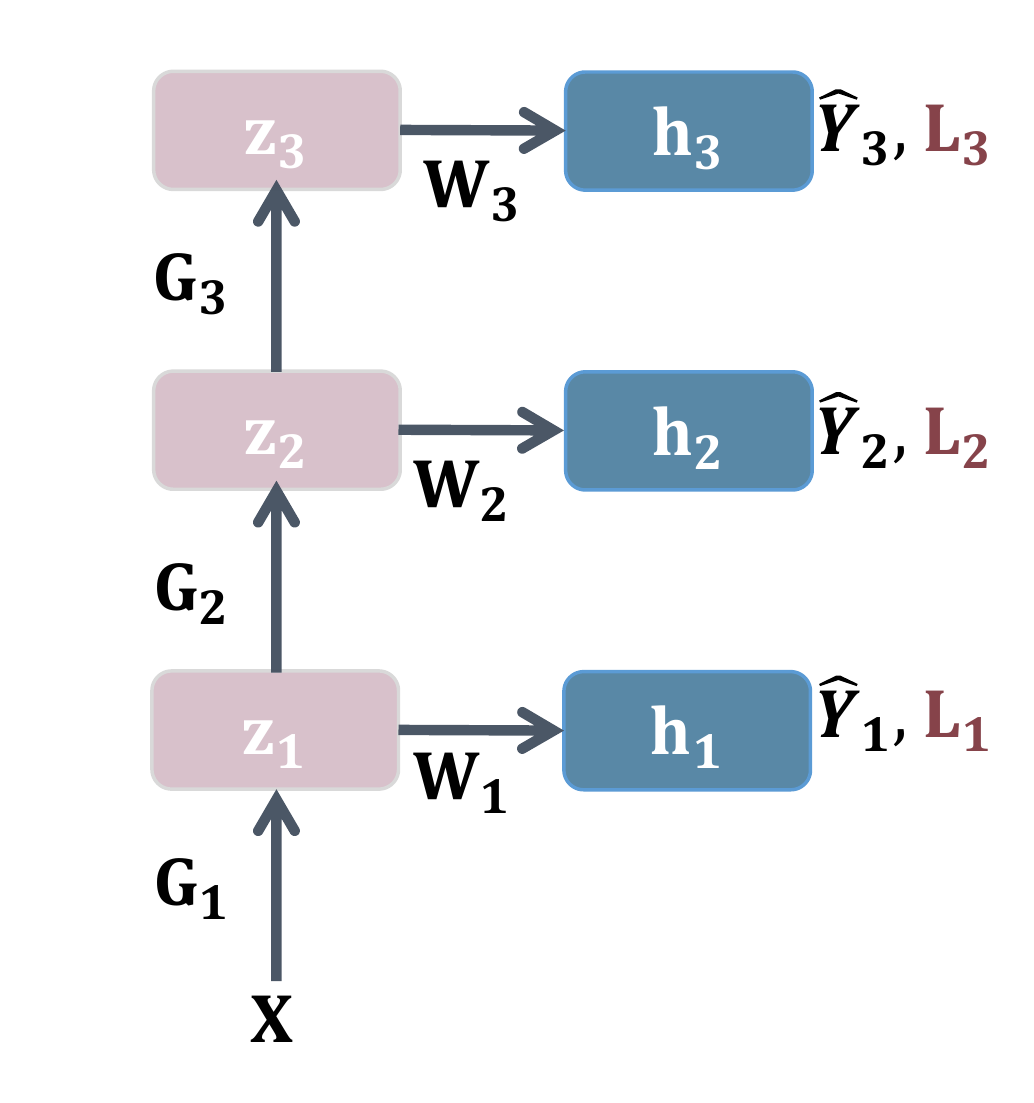}
			\end{minipage}
		}

	\end{center}
	\caption{Comparison of Backpropagation (BP) algorithm, four non-BP algorithms (FF, DFA, DRTP, PEPITA) and the proposed Cascaded Forward (CaFo) algorithm.}
	\label{fig:comparision}
\end{figure*}

\subsection{Relationship to Previous Non-BP Algorithms} 

In CaFo, the training mechanism of neural blocks shares similarities with DFA~\cite{nokland2016direct}. Nevertheless, each predictor independently makes its greedy decision about the class to which the input belongs, according to its local loss. This approach differs from DFA, which relies solely on a global loss. The predictors are trained on intermediate features extracted at various scales, enhancing the model performance by capturing more intricate patterns, as demonstrated in the experimental results depicted in Figure~\ref{fig:layers1}.

Similarly, DRTP~\cite{frenkel2021learning} and PEPITA~\cite{dellaferrera2022error} update each layer in a forward direction. However, DRTP updates layers directly using ground truth without considering global or local loss calculations, potentially limiting the learning of class distributions. PEPITA updates layers based on the difference in intermediate representations between two forward passes, which is also not directly tied to classification loss.

Our CaFo algorithm builds upon the FF algorithm~\cite{hinton2022forward}, and shares significant similarities with it on local optimization. 
They both have advantages over the traditional backpropagation methods, such as the ability to handle non-differentiable components and increased flexibility in network architecture. However, the differences between FF and CaFo are also significant. For FF, positive and negative samples are selected and samples are used to calculate a local loss for each layer. 
Once the local loss of the $i$-th layer is calculated (i.e., $\textbf{L}_{i}$), the weights of the $i$-th layer (i.e., $\textbf{W}_{i}$) will be updated according to $\textbf{L}_{i}$, and after that the $\textbf{W}_{i}$ will be fixed during the next training stage for the $(i+1)$-th layer. 
The loss for each layer is calculated in sequence, and the layers of FF are sequentially updated by greedy decision. 
In contrast, our CaFo has only one input stream, avoiding the negative sampling process. Although the sequential update of each block in our framework is similar to FF, the local loss in CaFo directly measures the error between the predicted multi-class distribution and ground truth, rather than a simple goodness estimation. Moreover, since the pre-trained neural blocks enable independent training of the attached predictor without dependencies on the training outcomes of previous blocks, our method not only enhances portability and flexibility compared with FF but also facilitates straightforward deployment into parallel acceleration systems when the blocks are randomly parameterized without training.

The graphical depiction and comparison of BP, non-BP algorithms and proposed CaFo are shown in Figure~\ref{fig:comparision}.


\subsection{Biological Plausibility}

By introducing the aforementioned training strtegies for neural blocks and predictors, our approach overcomes several of the biological unrealistic aspects associated with backpropagation. Firstly, CaFo avoids the weight transport problem since it does not utilize weight-specific feedback signals. Secondly, with regard to locality, the prescribed updates for each synapse are solely based on the activity of the connected nodes, avoiding any global errors that may perturb node activity. Thirdly, CaFo does not freeze the neural activity to propagate and apply the modulatory signal. Finally, in CaFo, updates are performed block by block in a sequential manner, partially mitigating the update locking issue. The weights of the first predictor can be updated right at the second forward pass, not requiring waiting for downstream predictor updates. This allows the forward computation for the next input sample at the first predictor to start in parallel with the update of the second predictor from the previous sample, suggesting that CaFo can be suitable for edge computing devices requiring fast processing of the input signals.

\subsection{Relationship to Bagging} 
Bagging (Bootstrap Aggregating)~\cite{breiman1996bagging} is an ensemble learning technique that combines multiple models to enhance the overall performance of a machine learning algorithm. The basic idea of bagging is to train some individual models on different subsets of the training data, and then combine their outputs by taking the average of majority vote. Bagging eliminates the variance of the individual models and prevents overfitting by reducing the impact of noisy data points or outliers. This is achieved by generating multiple random samples of the training data, each of which is adopted to train a different model. The results are then combined to make a final prediction.  

Our method can be viewed as a special case of Bagging, where the predictors are trained on intermediate features extracted at various scales, rather than on the original images. In other words, our approach does not train individual models on different subsets of the training data, but assigns each individual model with a specific intermediate representations of the complete training data, which has the advantage of utilizing the information contained in intermediate features extracted at different scales, thus helping to capture more complex patterns to improve the performance.

\subsection{Relationship to Co-training} 
Co-training~\cite{blum1998combining} is a popular semi-supervised learning method that trains multiple classifiers on the same dataset with different data views to improve classification accuracy by leveraging unlabeled data. The process involves initializing several classifiers with different feature sets and training them on labeled data subsets. The classifiers then exchange predictions on unlabeled data and retrain on new labeled data subsets using these predictions. This iterative process continues until convergence, with classifiers refining their feature sets and predictions. 

As a semi-supervised learning technique, co-training is usually applied in scenarios where the labeled data is scarce but there is a large amount of unlabeled data available. In contrast, our method is a fully-supervised approach that trains multiple predictors for each block in parallel, allowing for more efficient and stable optimization of the model in supervised learning scenarios. As our method and co-training share the same idea of training multiple modules to improve the overall performance, our approach can be seen as the first stage of a co-training process, and can be easily transformed into a co-training method if an augmentation strategy for training set is introduced.

\section{Experiments}

The aim of this paper is to introduce the CaFo algorithm and to show that it works in relatively small neural networks containing a few million connections. We evaluate the proposed algorithm on four benchmarks in comparison with four non-BP algorithms. Future research will focus on investigating the scalability of the proposed approach to handle large neural networks with a much greater number of connections.

\subsection{Datasets and Experimental Settings}\label{sec:datasets}

The experiments are conducted on  the MNIST~\cite{lecun1998gradient}, CIFAR-10~\cite{krizhevsky2009learning}, CIFAR-100~\cite{krizhevsky2009learning} and Mini-ImageNet~\cite{vinyals2016matching} datasets, all of which are widely-used in evaluating image classification algorithms. The standard splits of MNIST, CIFAR-10 and CIFAR-100 are directly used. In the case of Mini-ImageNet, which is often used for evaluating few-shot learning algorithms, all the classes are mixed, and for each class, 500 images are randomly sampled for training and 100 other images for testing.

We compare the proposed CaFo algorithm with four non-BP baseline algorithms: DFA~\cite{nokland2016direct}, DRTP~\cite{frenkel2021learning}, PEPITA~\cite{dellaferrera2022error}, and FF~\cite{hinton2022forward}. Specifically, for MNIST and CIFAR-10, we report the results of both the original version of FF~\cite{hinton2022forward} and our reproduced version. For CIFAR-100 and Mini-ImageNet, we only report the results of our reproduced version, as the source code of the original version is not publicly available. 

For the proposed CaFo algorithm, we set the number of blocks $r$=3. Each block consists of a convolutional layer followed by a ReLU activation function, a max-pooling layer and a batch normalization layer. The convolutional layer of the three blocks have the same kernel size, padding and stride, which are set to be $3 \times 3$, 1 and 1, respectively; and the output channels are set to be 32, 128, and 512 respectively.

\begin{table}[t]
	\renewcommand{\arraystretch}{1.35} 
	\caption{Error rate on MNIST and CIFAR-10. The best and the second best results on each dataset are indicated with \HL{red} and \hl{blue} colors, respectively.}\label{tab:smalldata}
	\begin{center}
		\begin{tabular}{lrrrr}
			\hline
			& \multicolumn{2}{c}{\textbf{MNIST}}                                                     & \multicolumn{2}{c}{\textbf{CIFAR-10}}                                                   \\ \hline
			\textbf{Error rate (\%)}& \multicolumn{1}{l}{Training} & \multicolumn{1}{l}{Test} & \multicolumn{1}{l}{Training} & \multicolumn{1}{l}{\ Test}\\ \hline
			\textbf{DFA}               &  0.77  \ \  \                                     & 1.37                               & 16.75    \ \ \                                 & 34.29     \ \                               \\
			\textbf{DRTP}               & 0.62 \ \ \                                      & 1.41                               & 32.55    \ \ \                                 & 39.50     \ \                               \\
			\textbf{PEPITA}               & 1.81 \ \ \                                      & 1.84                               & 12.14    \ \ \                                 & 36.55     \ \                               \\
			\textbf{FF (Hinton)}               & -  \quad\quad                                     & 1.37                              & 24.00   \ \ \                                 & 41.00    \ \    \\
			\textbf{FF (reprodu.)}               & 0.57   \ \  \                                   & 2.02                              & 10.36  \ \ \                                  & 46.03      \ \                                \\
			\hline
			\textbf{CaFo+MSE (rand)}             &    0.71      \ \ \                         &    1.55                           & 12.56    \ \ \                            & 34.79        \ \            \\ 
			\textbf{CaFo+CE (rand)} & 0.04       \ \ \                          & 1.30                              &     9.22  \ \ \                         & 32.57           \ \                 \\ 
			\textbf{CaFo+SL (rand)} & 0.06    \ \ \                    &           1.20                    &    9.58  \ \ \                            &      33.74  \ \                 \\     \hline
			\textbf{CaFo+MSE (dfa)} & 0.48    \ \ \                    &           1.24                    &    10.99  \ \ \                            &      33.46  \ \   \\
			
			\textbf{CaFo+CE (dfa)} & 0.07    \ \ \                    &           \hl{\textbf{1.18}}                    &    16.58  \ \ \                            &      \HL{\textbf{30.52}}   \ \  \\                 
			\textbf{CaFo+SL (dfa)} & 0.17    \ \ \                    &           \HL{\textbf{1.05}}                    &    11.42  \ \ \                            &      \hl{\textbf{31.51}}  \ \   \\
			\hline
		\end{tabular}
	\end{center}
\end{table}


\begin{table}[t]
	\renewcommand{\arraystretch}{1.35} 
	\caption{Error rate on CIFAR-100 and Mini-ImageNet}\label{tab:largedata}
	\begin{center}
		\begin{tabular}{lrrrr}
			\hline
			& \multicolumn{2}{c}{\textbf{CIFAR-100}}                                                     & \multicolumn{2}{c}{\textbf{Mini-ImageNet}}   \ \ \                                               \\ \hline
			\textbf{Error rate (\%)}& \multicolumn{1}{l}{Training} & \multicolumn{1}{l}{\ Test} & \multicolumn{1}{l}{\ Training} & \multicolumn{1}{l}{\ \ \ Test}\\ \hline
			\textbf{DFA}               & 8.93 \ \ \ \                                      & 62.02                               & 4.87    \ \ \                                 & 84.61     \ \                               \\
			\textbf{DRTP}               & 12.07 \ \ \ \                                      & 63.15                               & 80.51    \ \ \                                 & 87.91     \ \                               \\
			\textbf{PEPITA}               & 1.32 \ \ \ \                                      & 61.26                               & 0.16    \ \ \                                 & 90.87     \ \                               \\   
			\textbf{FF (reprodu.)}               & 29.32  \ \ \ \                                    & 78.67                              &  27.84 \ \ \                                  &    91.71   \ \                                \\
			\hline
			\textbf{CaFo+MSE (rand)}             & 6.06         \ \ \ \                       & 63.94                              &  6.66    \ \ \                            &  85.50       \ \            \\ 
			\textbf{CaFo+CE (rand)} & 0.06      \ \ \ \                          & 59.24                              &   0.01   \ \ \                        &       \textbf{\hl{78.20}}      \ \                 \\ 
			\textbf{CaFo+SL (rand)} &     0.02 \ \ \ \                            &         61.96                      &    0.02  \ \ \                            &      81.47  \ \                 \\  \hline
			\textbf{CaFo+MSE (dfa)} & 5.59    \ \ \ \                   &           63.30                    &    7.57  \ \ \                            &      83.08  \ \   \\
			\textbf{CaFo+CE (dfa)} & 2.72   \ \ \ \                    &           \textbf{\HL{57.87}}                    &    0.01  \ \ \                            &      \textbf{\HL{77.59}}  \ \                 \\ 
			\textbf{CaFo+SL (dfa)} & 0.29    \ \ \ \                   &           \hl{\textbf{58.97}}                    &    0.02  \ \ \                            &      79.55  \ \   \\
			\hline
		\end{tabular}
	\end{center}
\end{table}

As described in the previous section, different error measure functions can be adopted in our framework to guide the training. For analysis of the affects of different measure functions, in this section we report the results of three versions of our CaFo algorithms that adopt MSE (CaFo+MSE), cross-entropy (CaFo+CE) and sparsemax loss (CaFo+SL) as the measure function respectively. The predictor is a fully connected layer without bias, followed by a softmax layer for CE and SL. For CaFo+CE and CaFo+SL, each predictor is trained for 5000 epochs on MNIST and CIFAR-10, and for 1000 epochs on CIFAR-100 and Mini-ImageNet. We pre-initialize the parameters of each block with Kaiming uniform~\cite{he2015delving} (abbr. rand) and the pre-trained weights of DFA (abbr. dfa). All experiments are run on AMD EPYC 7542 32-Core Processor with an Nvidia GeForce RTX 3090 GPU.

\subsection{Performance Comparison}

The comparison of these algorithms on four datasets is summarized in Tables~\ref{tab:smalldata} and \ref{tab:largedata}, in which the best and the second best results on each dataset are indicated with red and blue colors, respectively. It is obvious our method achieves overall improvements in test error rate compared with the baseline algorithms, obtaining the best or second best results on all datasets.

Even the simplest version (CaFo+MSE) demonstrates comparable test results with DFA, DRTP and PEPITA, while outperforming the reproduced FF on all datasets. For FF, we observe a significant discrepancy in its results between the two tables. FF has low training error when the dataset is easy to discriminate (e.g., CIFAR-10 and MNIST), but has high training error when tackling much more complex datasets (e.g., CIFAR-100 and Mini-ImageNet). The reason for the underfitting of FF probably lies in that the contrastive loss function based on positive and negative samples may not provide effective guidance. If the training samples for each class are scarce (e.g., CIFAR-10 vs. CIFAR-100) or the samples contain more irrelevant information (e.g., CIFAR-10 vs. Mini-ImageNet), the goodness estimation of positive and negative samples may not effectively help it to learn the potential distribution for each class because of the uncertain quality of negative samples and the lack of supervision information directly related to the labels. This may lead to the poor fitting for the training set. Similarly, the PEPITA algorithm that does not use classification loss as updating guidance also experiences consistently high training errors.

In contrast, our method exhibits a smaller discrepancy between these datasets and demonstrates better fitting ability in general. For test error rate,  our CaFo outperforms all baseline algorithms by remarkable margins, especially on more complex datasets. For fair comparison, the results of CaFo are reported without deployment of any regularization trick, and thus the performance still has the potential to be improved by means of regularization strategies.

\begin{table*}[t]
	\renewcommand{\arraystretch}{1.35} 
	\caption{Running time (seconds) for training and testing}\label{tab:time}
	\begin{center}
		\begin{tabular}{lrrrrrrrr}
			\hline
			& \multicolumn{2}{c}{\textbf{MNIST}}                                                     & \multicolumn{2}{c}{\textbf{CIFAR-10}}     &
			\multicolumn{2}{c}{\textbf{CIFAR-100}}&
			\multicolumn{2}{c}{\textbf{Mini-ImageNet}}                                              \\ \hline
			\textbf{Time (seconds)}& \multicolumn{1}{l}{Training} & \multicolumn{1}{l}{Test} & \multicolumn{1}{l}{Training} & \multicolumn{1}{l}{Test}  & \multicolumn{1}{l}{Training} & \multicolumn{1}{l}{Test}   & \multicolumn{1}{l}{Training} & \multicolumn{1}{l}{Test}\\ \hline
			\textbf{FF (reprodu.)}               & 1655.22  \ \ \                                     & 0.02                              & 2374.30  \ \ \                                  & 0.04      \           & 2376.77  \ \ \ & 0.42 & 3668.35 \ \ \ &  2.03    \                 \\
			\textbf{DFA }               & 925.48  \ \ \                                     & 0.05                              &  956.06 \ \ \                                  & 0.08      \           & 891.40  \ \ \ & 0.07 & 12942.45 \ \ \ &  0.31    \                 \\
			\textbf{DRTP }               & 1054.29  \ \ \                                     & 0.06                              & 955.01  \ \ \                                  & 0.08      \           & 892.88  \ \ \ & 0.08 & 12966.59 \ \ \ &  0.31    \                 \\
			\textbf{PEPITA }               & 1551.94  \ \ \                                     & 0.10                              & 1725.66  \ \ \                                  & 0.11      \           & 1717.39  \ \ \ & 0.11 & 41522.68 \ \ \ &  0.32    \                 \\
			\hline
			\textbf{CaFo+MSE (rand)}             & 0.62         \ \ \                          & 0.06                              & 1.21    \ \ \                            & 0.07        \      & 1.40 \ \ \ & 0.08 & 22.55 \ \ \ & 0.32  \    \\ \hline
			\textbf{CaFo+CE (rand)} &  612.22  \ \ \                               & 0.05                              & 873.56      \  \ \                        & 0.07           \     & 1178.42 \ \ \ & 0.08 & 8101.42 \ \ \ & 0.52  \           \\ 
			\textbf{CaFo+SL (rand)} & 837.56   \ \ \                      &          0.06                    &    1082.45  \ \ \                            &     0.09   \        & 2192.56 \ \ \ & 0.10 & 14128.06  \ \ \ & 0.49    \     \\    \hline
			\textbf{CaFo+CE (dfa)} & 1504.29   \ \ \                      &          0.05                    &    1770.82  \ \ \                            &     0.07   \        & 2059.08 \ \ \ & 0.08 & 20842.13  \ \ \ & 0.52    \     \\
			\hline
		\end{tabular}
	\end{center}
\end{table*}

When comparing the three variants of our method that have randomly initialized block parameters, we observe that they obtain overall similar performance. CaFo+MSE has the highest test and train error on all the datasets, which suggests that MSE may not be good enough for error estimation in comparison with other two variants that adopt classification-oriented loss. In contrast, both CaFo+CE and CaFo+SL exhibit low training and test errors on all the four datasets, indicating the effectiveness of gradient descent optimization strategy and the superiority of CE and SL loss. 

Given DFA's effectiveness as a non-BP approach, we employ this method to train the parameters of each block and incorporate them into CaFo. The results show that CaFo (dfa) stands out as the better promising variant in comparision with CaFo (rand), achieving the best performance on all datasets. This intriguing observation suggests the feasibility of integrating well-learned blocks trained by non-BP approaches into CaFo, potentially enhancing the representation ability of our model.

\subsection{Time Comparison}
Furthermore, we report the time required for training and testing our method as shown in Table~\ref{tab:time}. For each trial, we report the time required for the entire training process and the test time that includes the time for the forward pass of the test samples and the time for calculating the predicted category. The experimental settings are the same as those in Tables~\ref{tab:smalldata} and ~\ref{tab:largedata}.

Comparing our three variants with baseline algorithms with whose blocks are randomly initialized, we observe that CaFo+MSE exhibits significant improvements in training efficiency, which has a much shorter training time than other reported approaches due to its direct calculation of the close-form solution without iterative training. Additionally, CaFo+CE and CaFo+SL also demonstrate comparable training efficiency when compared with baseline algorithms. Generally, our method shows significant improvements in training efficiency on MNIST, CIFAR-10, while it requires more time on CIFAR-10 and Mini-ImageNet. This is because we divide the training set of CIFAR-100 and Mini-ImageNet into 20 and 200 training batches, respectively, for the latter two variants due to memory limitation, which leads to the extra time consumption of data movement. However in the trials of MNIST and CIFAR-10 where the batch number equals one, we find the two CaFo variants are trained significantly faster than baseline algorithms. CaFo+CE (dfa) requires more time than other variants due to its time cost, which includes both the pre-initialization time of blocks and the training time of predictors. Despite this, the overall time cost is still acceptable when compared with baselines.

\begin{table*}[t]
	\renewcommand{\arraystretch}{1.35} 
	\caption{Error rate with different training strategies for neural block}\label{tab:training_strategy}
	\begin{center}
		\begin{tabular}{lrrrrrrrr}
			\hline
			& \multicolumn{2}{c}{\textbf{MNIST}}                                                     & \multicolumn{2}{c}{\textbf{CIFAR-10}}     &
			\multicolumn{2}{c}{\textbf{CIFAR-100}}&
			\multicolumn{2}{c}{\textbf{Mini-ImageNet}}                                              \\ \hline
			\textbf{Error rate (\%)}& \multicolumn{1}{l}{Training} & \multicolumn{1}{l}{Test} & \multicolumn{1}{l}{Training} & \multicolumn{1}{l}{Test}  & \multicolumn{1}{l}{Training} & \multicolumn{1}{l}{Test}   & \multicolumn{1}{l}{Training} & \multicolumn{1}{l}{Test}\\ \hline
			\textbf{CaFo+MSE (rand)}               & 0.71  \ \ \                                     & 1.55                              &  12.56 \ \ \                                  & 34.79      \           & 6.06  \ \ \ & 63.94 & 6.66 \ \ \ & 85.50    \                 \\
			\textbf{CaFo+CE (rand)}               & 0.04  \ \ \                                     & 1.30                              & 9.22  \ \ \                                  & 32.57      \           & 0.06  \ \ \ & 59.24 & 0.01 \ \ \ &  78.20    \                 \\
			\textbf{CaFo+SL (rand) }               & 0.06  \ \ \                                     & 1.20                              & 9.58  \ \ \                                  & 33.74      \           & 0.02  \ \ \ & 61.96 & 0.02 \ \ \ &  81.47    \                 \\
			\hline
			\textbf{CaFo+MSE (dfa)}             & 0.48         \ \ \                          & 1.24                              & 10.99    \ \ \                            & 33.46        \      & 5.59 \ \ \ & 63.30 & 7.57 \ \ \ & 83.08  \    \\ 
			\textbf{CaFo+CE (dfa)}             & 0.07         \ \ \                          & 1.18                              & 16.58    \ \ \                            & 30.52        \      & 2.72 \ \ \ & 57.87 & 0.01 \ \ \ & 77.59  \    \\
			\textbf{CaFo+SL (dfa)}             & 0.17         \ \ \                          & 1.05                              & 11.42    \ \ \                            & 31.51        \      & 0.29 \ \ \ & 58.97 & 0.02 \ \ \ & 79.55  \    \\  \hline
			\textbf{CaFo+MSE (drtp)} &  1.23  \ \ \                               & 2.18                              & 69.05      \  \ \                        & 76.09           \     & 91.66 \ \ \ & 96.64 & 91.08 \ \ \ & 97.69  \           \\ 
			\textbf{CaFo+CE (drtp)} & 1.33   \ \ \                      &          1.64                    &    30.40  \ \ \                            &     37.58   \        & 23.82 \ \ \ & 64.50 & 0.03  \ \ \ & 81.92    \     \\ 
			\textbf{CaFo+SL (drtp)} & 0.64   \ \ \                      &          1.18                    &    36.91  \ \ \                            &     43.98   \        & 29.62 \ \ \ & 64.13 & 0.02  \ \ \ & 83.32    \     \\   \hline
			\textbf{CaFo+MSE (bp)} & 0.51   \ \ \                      &          1.74                    &    1.98  \ \ \                            &     29.52   \        & 2.41 \ \ \ & 62.51 & 6.41  \ \ \ & 82.18    \     \\
			\textbf{CaFo+CE (bp)} & 0.17   \ \ \                      &          1.06                    &    1.74  \ \ \                            &     26.30   \        & 0.03 \ \ \ & 56.52 & 0.01  \ \ \ & 74.95    \     \\
			\textbf{CaFo+SL (bp)} & 0.21   \ \ \                      &          0.96                    &    0.33  \ \ \                            &     27.17   \        & 0.02 \ \ \ & 58.60 & 0.01  \ \ \ & 76.58   \     \\
			\hline
			\textbf{BP} & 0.00   \ \ \                      &          0.77                    &    0.00  \ \ \                            &     24.36   \        & 0.02 \ \ \ & 54.11 & 0.05  \ \ \ & 73.76    \     \\
			\hline
		\end{tabular}
	\end{center}
\end{table*}

\begin{figure}[t!]
	\begin{center}
		
		\subfigure[MNIST]{
			\begin{minipage}[b]{0.45\textwidth}
				\centering
				\includegraphics[width=1\textwidth]{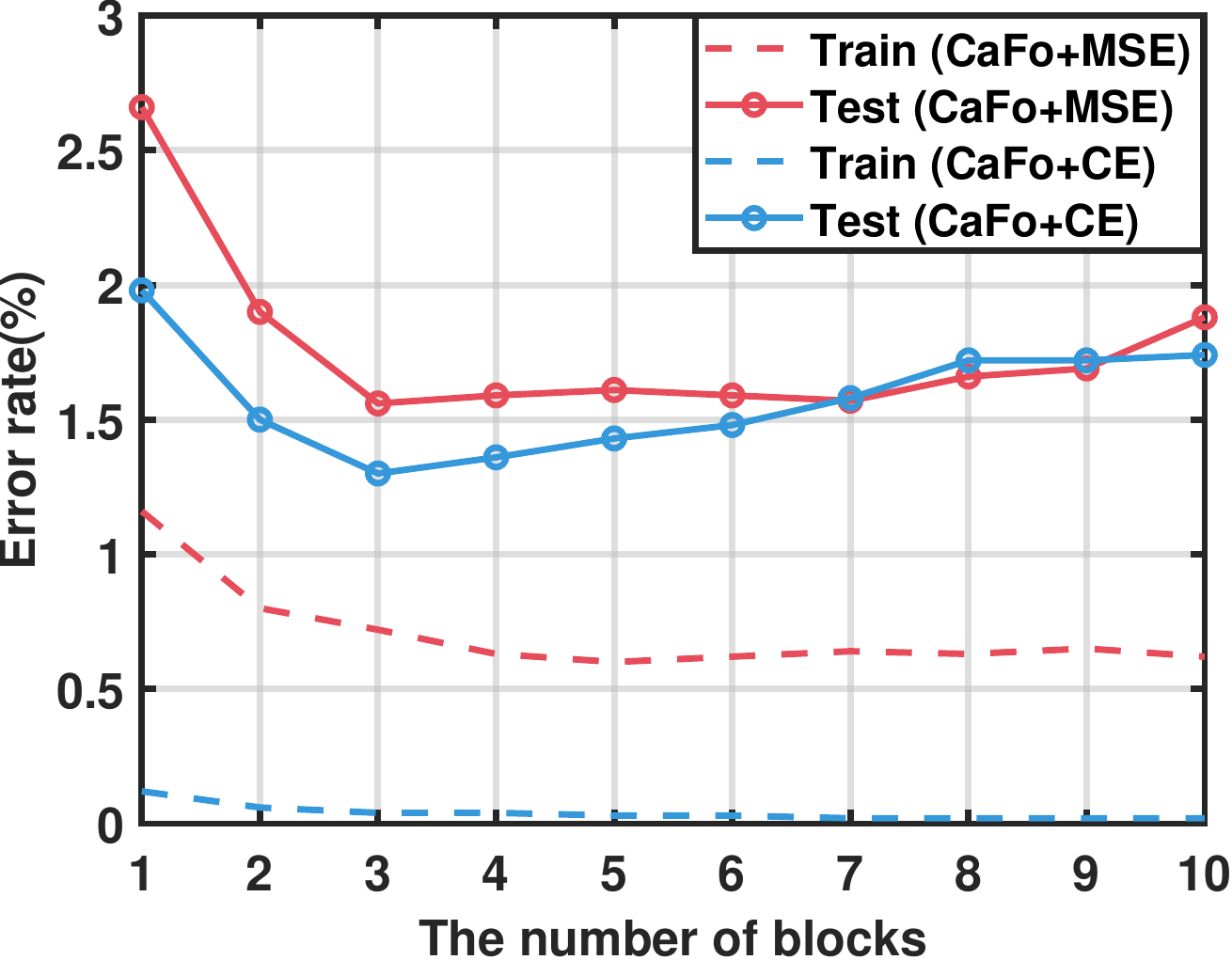}
			\end{minipage}
		}
		\subfigure[CIFAR-10]{
			\begin{minipage}[b]{0.45\textwidth}
				\centering
				\includegraphics[width=1\textwidth]{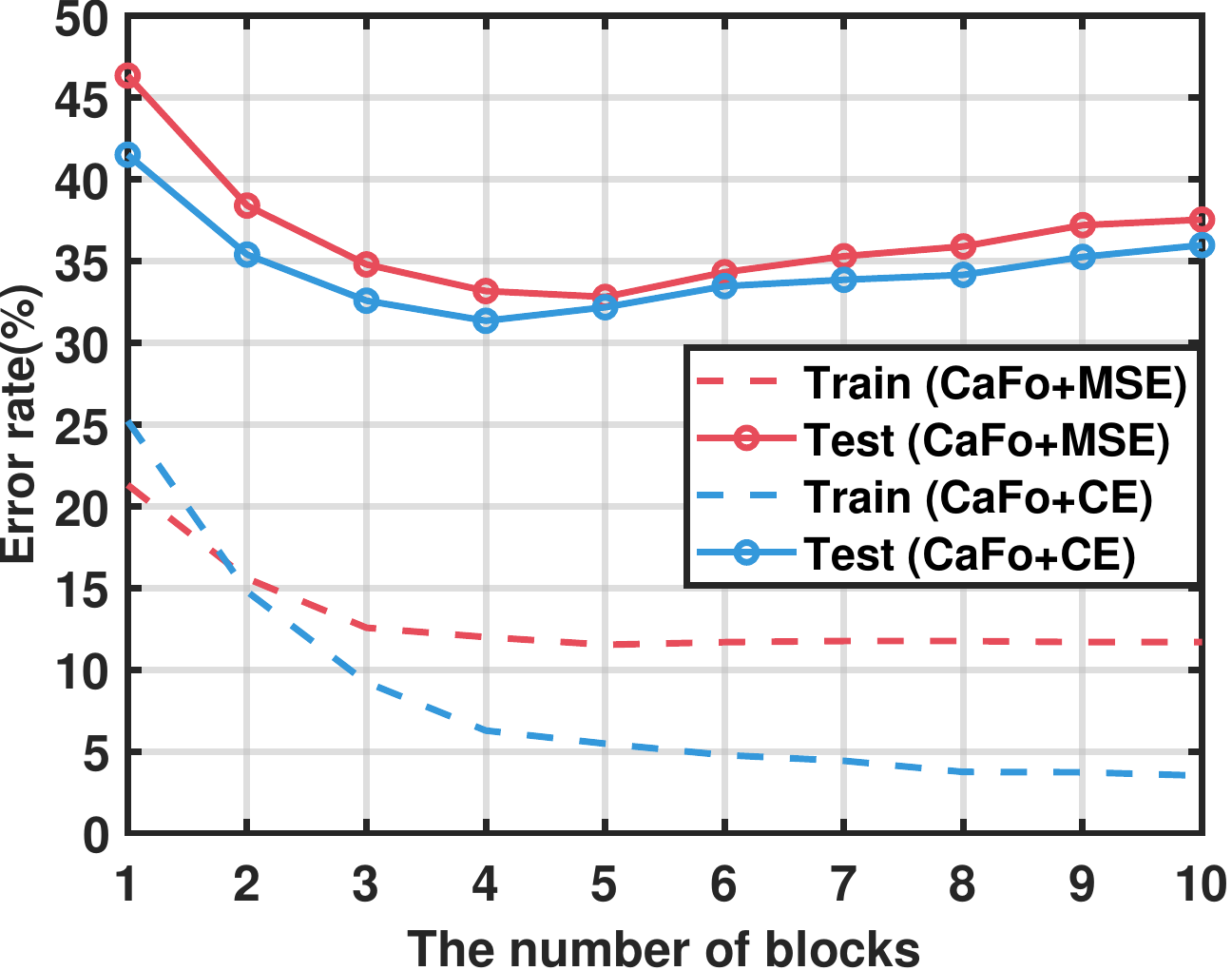}
			\end{minipage}
		}

	\end{center}
	\caption{Error rates with different numbers of blocks.}
	\label{fig:layers}
\end{figure}


In terms of test time, our methods demonstrate overall comparable efficiency in comparison with DFA, DRTP and PEPITA due to their similar testing mechanism. Compared with FF, our method has comparable efficiency on MNIST and CIFAR-10, but significantly faster test time on CIFAR-100 and Mini-ImageNet. The reason for this lies in that our methods directly output the prediction of multi-class distribution, rather than compute the goodness estimation for each candidate label as that in FF. It tremendously improves the test efficiency when the number of categories is relatively large, such as in CIFAR-100 and Mini-ImageNet.

\subsection{Investigation of different training strategies for neural block}

In Table~\ref{tab:training_strategy}, we present the error rates of CaFo under various neural block training strategies: random initialization (abbr. rand), backpropagation (abbr. bp), direct feedback alignment (abbr. dfa), and direct random target projection (abbr. drtp). Additionally, we include results for an end-to-end BP method (abbr. BP) using the same architecture.

CaFo trained with DFA displays an overall improvement compared with DRTP, highlighting DFA's effectiveness as a non-BP method for training neural blocks. Surprisingly, even trials with random initialization show better performance than DRTP trials, underscoring the negative impact of an inappropriate training strategy on our method's effectiveness.

In comparison with BP, although CaFo demonstrates superiority over competitive non-BP methods as evident in Tables~\ref{tab:smalldata} and \ref{tab:largedata}, a notable performance gap still exists. When neural blocks are trained via backpropagation, CaFo, while not classified as a non-BP method, exhibits better sample fitting and a smaller error rate. Thus, achieving results comparable to backpropagation in non-BP environments remains a crucial research concern for CaFo, necessitating further investigation.

\subsection{Investigation of the Number of Blocks}

In Tables~\ref{tab:smalldata} and \ref{tab:largedata} we fix the number of blocks $r$=3 for fair comparison with \cite{hinton2022forward}, in which the results of BP and FF with three layers are reported. However, the number of blocks may have diverse effects on different datasets. Here we investigate how the number of blocks influences the performance of our method. 

In Figure~\ref{fig:layers}  the error rates of  CaFo+MSE and CaFo+CE  with the number of blocks ranging from 1 to 10 are reported. As CIFAR-10 allows for a maximum of five $2 \times 2$ max-pooling layers with $stride$=2, and four are available for MNIST, we remove the max-pooling layer from some blocks to keep the number of max-pooling layers within the upper bound if necessary. We observe that the training error rates (dashed lines) continuously decrease with the increasing number of blocks, indicating an enhancement of fitting ability as more learnable predictors are introduced. However, the test error curves (solid lines) show that the model is more or less affected by overfitting when the number of blocks is large.

\begin{figure*}[t!]
	\begin{center}
		
		\subfigure[CaFo+MSE (rand)]{
			\begin{minipage}[b]{0.31\textwidth}
				\centering
				\includegraphics[width=1\textwidth]{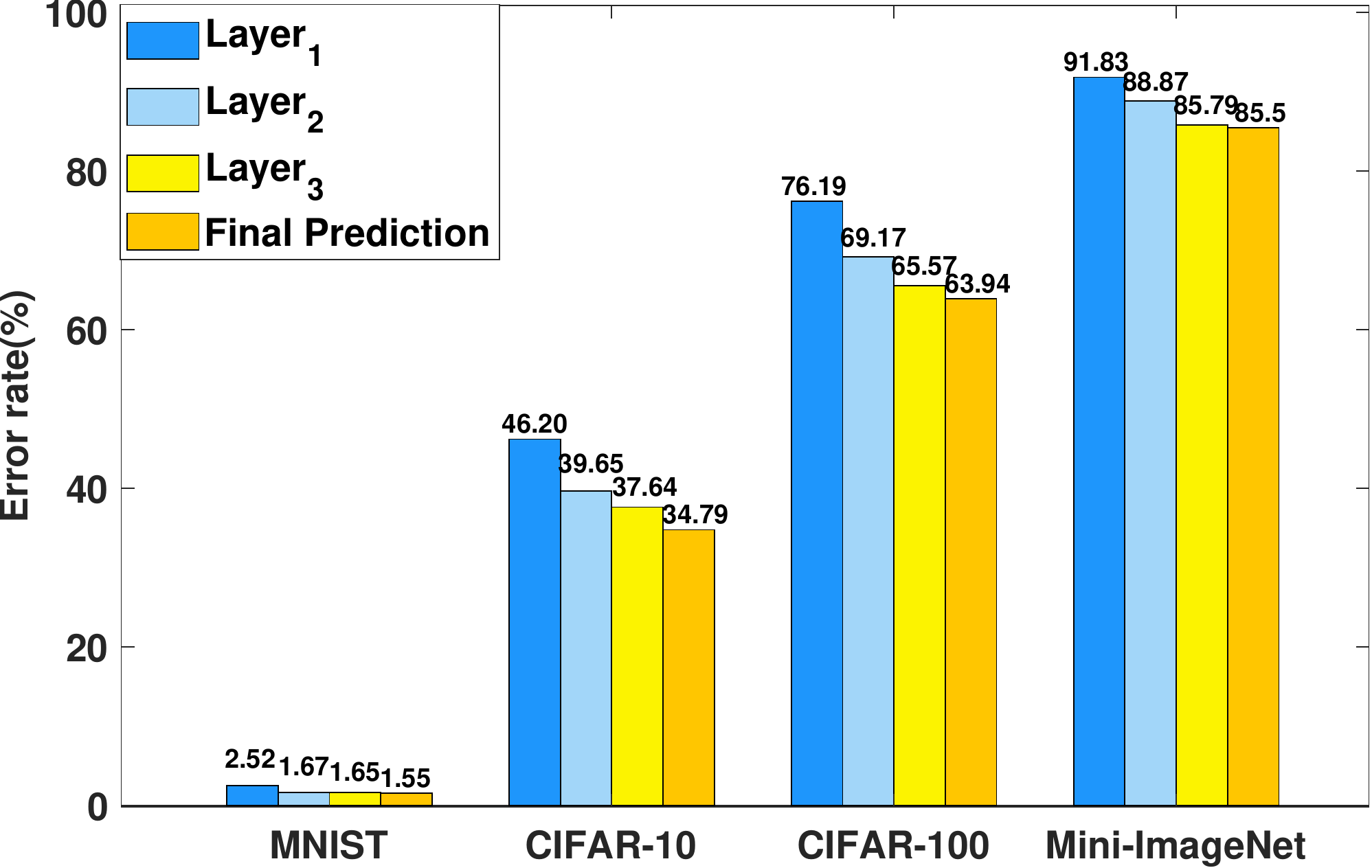}
			\end{minipage}
		}
		\subfigure[CaFo+CE (rand)]{
			\begin{minipage}[b]{0.31\textwidth}
				\centering
				\includegraphics[width=1\textwidth]{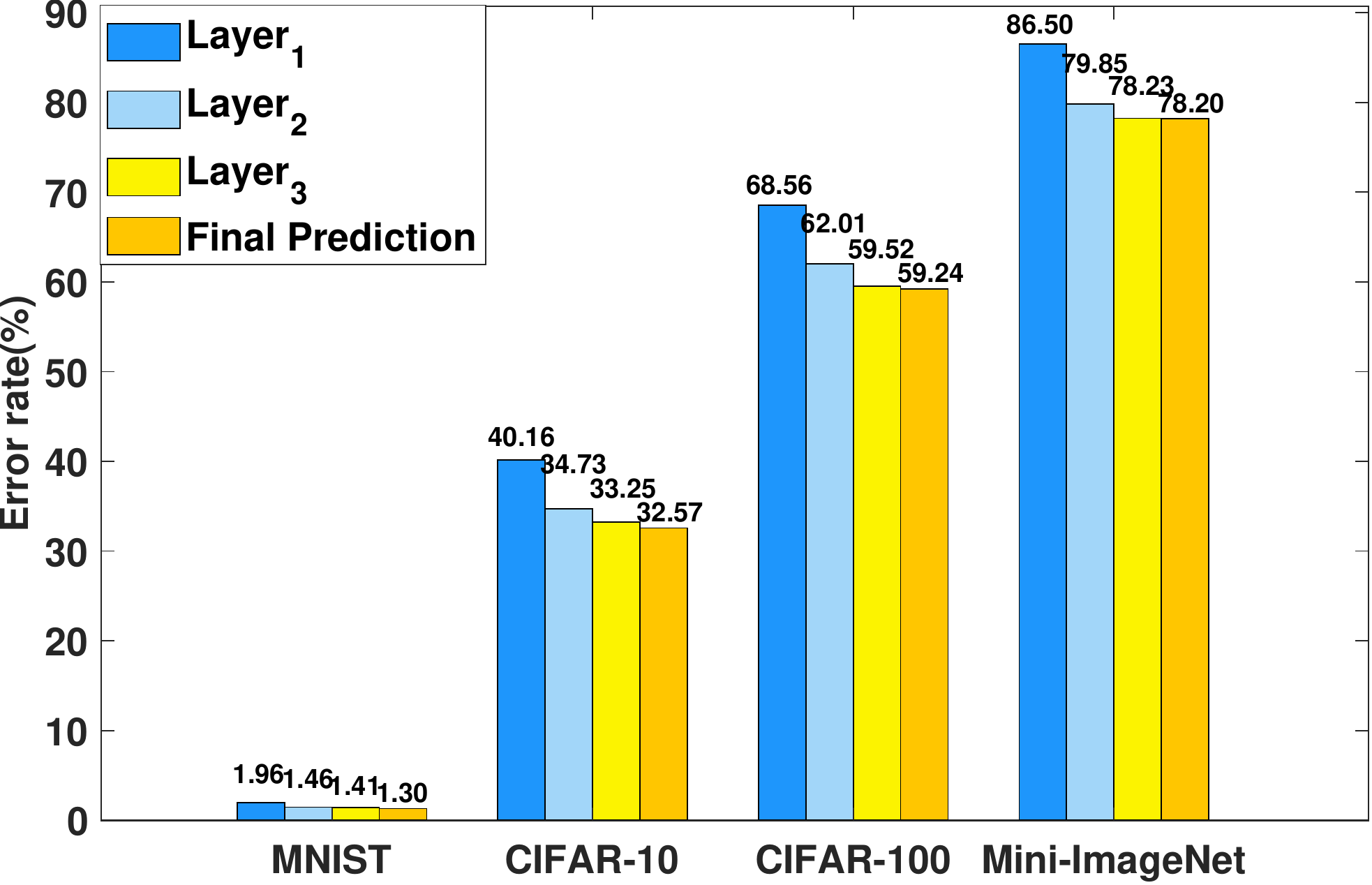}
			\end{minipage}
		}
		\subfigure[CaFo+SL (rand)]{
			\begin{minipage}[b]{0.31\textwidth}
				\centering
				\includegraphics[width=1\textwidth]{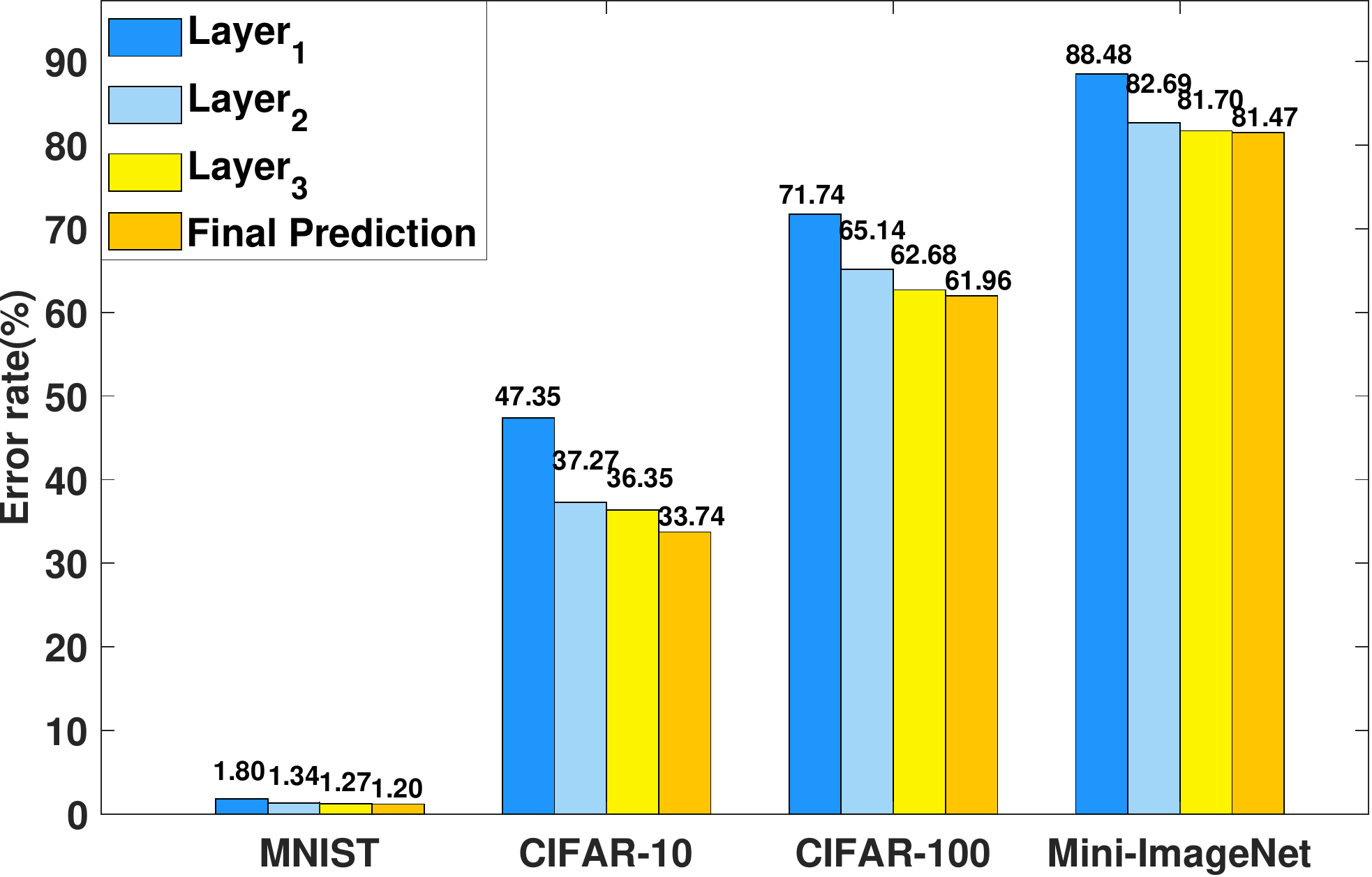}
			\end{minipage}
		}
		
		\subfigure[CaFo+MSE (dfa)]{
			\begin{minipage}[b]{0.31\textwidth}
				\centering
				\includegraphics[width=1\textwidth]{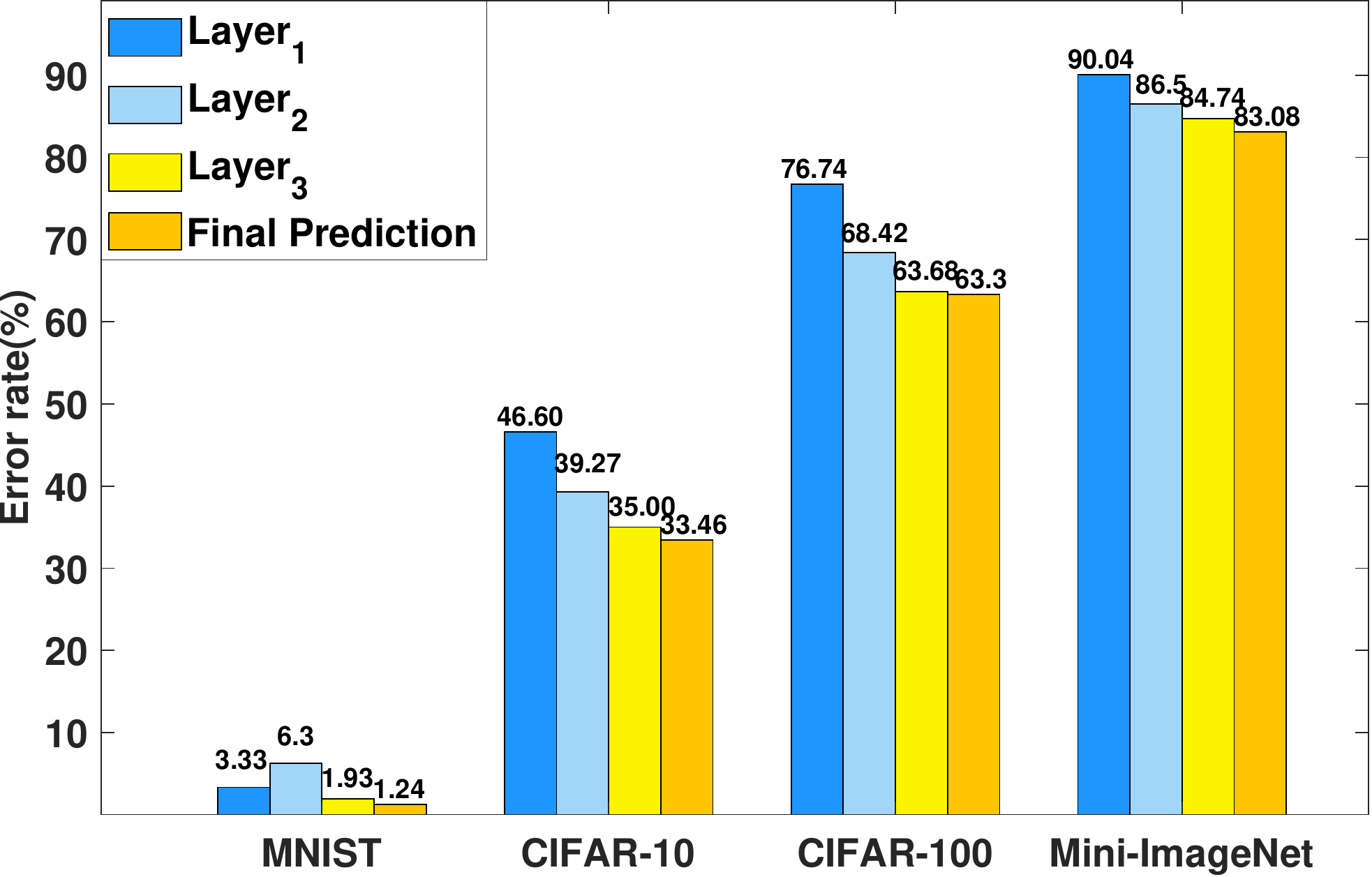}
			\end{minipage}
		}
		\subfigure[CaFo+CE (dfa)]{
			\begin{minipage}[b]{0.31\textwidth}
				\centering
				\includegraphics[width=1\textwidth]{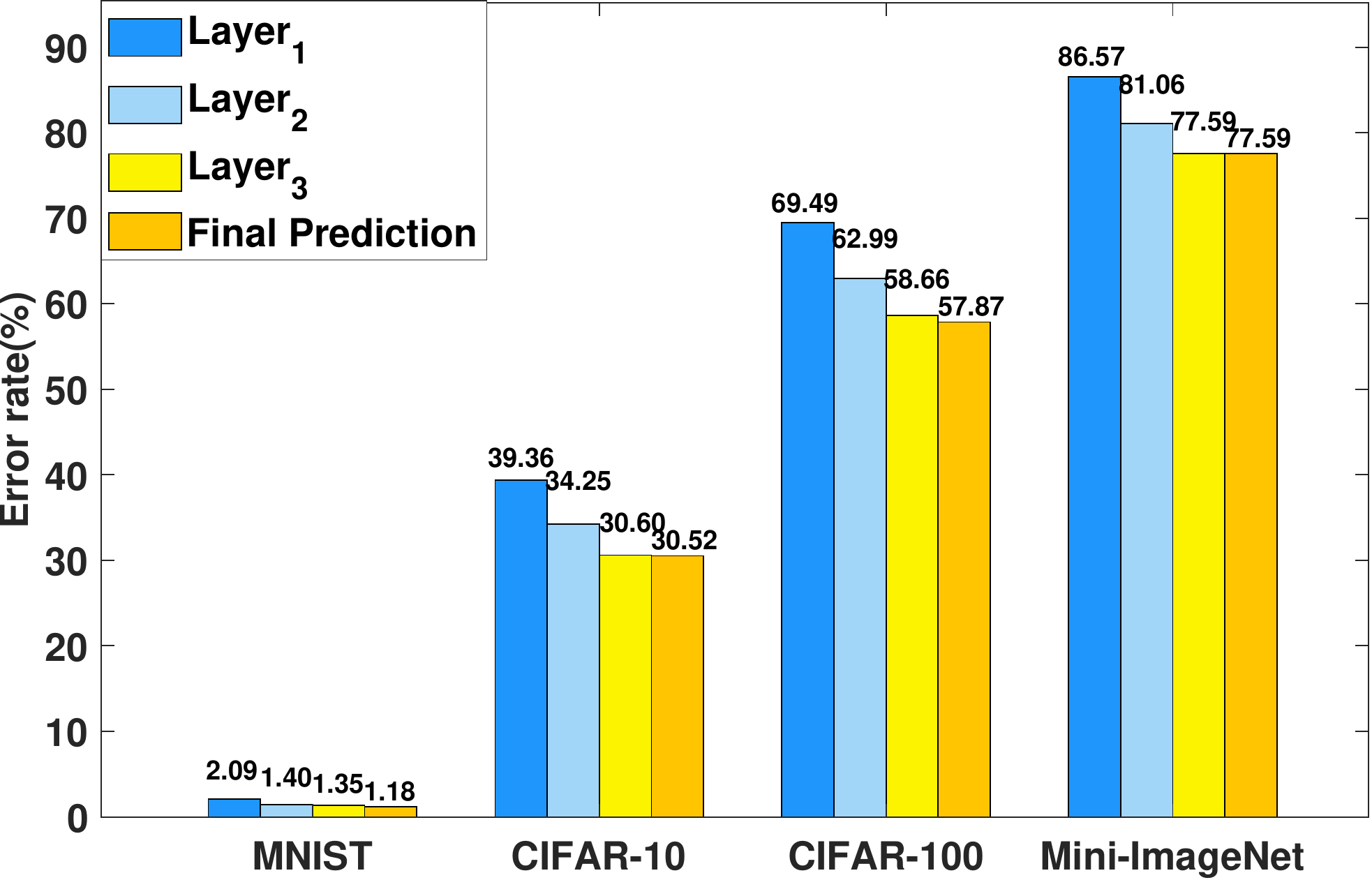}
			\end{minipage}
		}
		\subfigure[CaFo+SL (dfa)]{
			\begin{minipage}[b]{0.31\textwidth}
				\centering
				\includegraphics[width=1\textwidth]{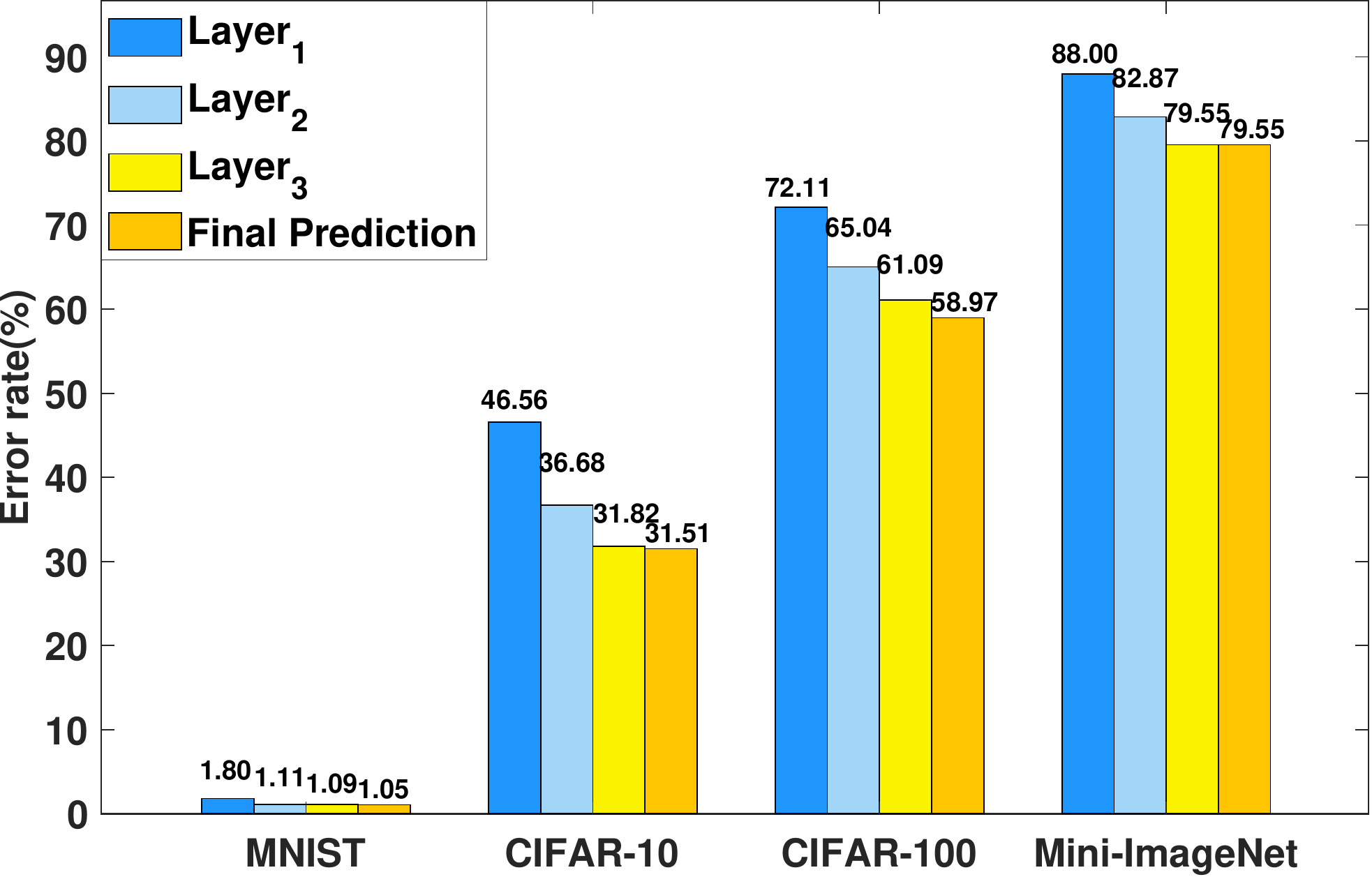}
			\end{minipage}
		}
	\end{center}
	\caption{Test error rate of each predictor.}
	\label{fig:layers1}
\end{figure*}

\subsection{Performance of each Predictor}

To investigate the performance of each predictor, we report the test error rate of each predictor for CaFo+MSE and CaFo+CE, on four datasets. As shown in Figure~\ref{fig:layers1}, the test error of the final prediction is lower than those of previous individual predictors. Moreover, we find that a deep-layer predictor tends to obtain better performance than a shallow-layer one, demonstrating that the deep features play a more important role in enabling the predictor to make correct classifications. Overall, the results in Figure~\ref{fig:layers1} well validate that the combination of all the predictors effectively contributes to the final prediction. Although both FF and our method illustrate that the straightforward sum of each block's prediction (goodness) is sufficient for comparable performance, however, designing more effective fusion strategies for these non-backpropagation approaches remains to be studied. 

\section{Conclusion}

In this paper, we propose a new learning framework, named CaFo, for deep neural networks that provides a viable alternative to the backpropagation algorithm. We show that CaFo has the feasibility of using only non-backpropagation strategies without the need for backward pass in the training process of neural blocks and predictors, which offers significant improvements in prediction accuracy, training efficiency and simplicity compared with previous algorithms. Extensive experiments clearly validate the superior performance over the competitive methods on several public benchmarks.

While our proposed method has demonstrated remarkable performance, several challenges persist, particularly regarding scalability to large-scale datasets and more complex network architectures, which are common hurdles in neural network learning. In our future endeavors, we intend to investigate approaches to extend CaFo's applicability to handle larger datasets and diverse architectures, such as graph neural networks (GNNs) and transformers. Furthermore, enhancing fusion strategies for CaFo is an area we plan to delve into, aiming to optimize the utilization of predictions from each predictor for improved overall performance.

\bibliographystyle{unsrt}  
\bibliography{references}

\end{document}